\documentclass[journal]{IEEEtran}
\usepackage{amsmath,amsfonts}
\usepackage{algorithmic}
\usepackage{array}
\usepackage[caption=false,font=normalsize,labelfont=sf,textfont=sf]{subfig}
\usepackage{textcomp}
\usepackage{stfloats}
\usepackage{url}
\usepackage{hyperref}
\usepackage{verbatim}
\usepackage{graphicx}
\usepackage{cite}
\usepackage{multirow}
\usepackage{makecell}
\usepackage{booktabs}
\usepackage{tabularx}
\usepackage{amssymb}
\usepackage{ulem}
\usepackage{epstopdf}
\usepackage[table, dvipsnames]{xcolor}
\usepackage[ruled,linesnumbered]{algorithm2e}
\def\BibTeX{{\rm B\kern-.05em{\sc i\kern-.025em b}\kern-.08em
    T\kern-.1667em\lower.7ex\hbox{E}\kern-.125emX}}
\usepackage{balance}

\begin{document}
\normalem

\title{Evolving Generalizable Parallel Algorithm Portfolios for Binary Optimization Problems via Domain-Agnostic Instance Generation}
\author{Zhiyuan~Wang,~\IEEEmembership{Student Member,~IEEE},
Shengcai Liu,~\IEEEmembership{Member,~IEEE},
\\
Peng Yang,~\IEEEmembership{Senior Member,~IEEE},
\and Ke Tang,~\IEEEmembership{Fellow,~IEEE}
\thanks{Zhiyuan Wang, Shengcai Liu, Peng Yang, and Ke Tang are with the Guangdong Provincial Key Laboratory of Brain-inspired Intelligent Computation, Department of Computer Science and Engineering, Southern University of Science and Technology, Shenzhen 518055, China.
Peng Yang is also with Department of Statistics and Data Science, Southern University of Science and Technology, Shenzhen 518055, China (email: wangzy2020@mail.sustech.edu.cn; liusc3@sustech.edu.cn; yangp@sustech.edu.cn; tangk3@sustech.edu.cn).

Corresponding authors: Shengcai Liu and Ke Tang.
}
}

\markboth{IEEE TRANSACTIONS ON EVOLUTIONARY COMPUTATION,~Vol.~XX, No.~X, XX~XXXX}{Evolving Generalizable Parallel Algorithm Portfolios for Binary Optimization Problems via Domain-Agnostic Instance Generation}

\maketitle

\begin{abstract}
Generalization is the core objective when training optimizers from data.
However, limited training instances often constrain the generalization capability of the trained optimizers.
Co-evolutionary approaches address this challenge by simultaneously evolving a parallel algorithm portfolio (PAP) and an instance population to eventually obtain PAPs with good generalization.
Yet, when applied to a specific problem class, these approaches have a major limitation.
They require practitioners to provide instance generators specially tailored to the problem class, which is often non-trivial to design.
This work proposes a general-purpose, off-the-shelf PAP construction approach, named domain-agnostic co-evolution of parameterized search (DACE), for binary optimization problems where decision variables take values of 0 or 1.
The key novelty of DACE lies in its neural network-based domain-agnostic instance representation and generation mechanism that eliminates the need for domain-specific instance generators.
The strong generality of DACE is validated across three real-world binary optimization problems: the complementary influence maximization problem (CIMP), the compiler arguments optimization problem (CAOP), and the contamination control problem (CCP).
Given only a small set of training instances from these problem classes, DACE, without requiring domain knowledge, constructs PAPs with even better generalization performance than existing approaches on all three classes, despite their use of domain-specific instance generators.
\end{abstract}


\section{Introduction}

\IEEEPARstart{I}{n} recent decades, search-based methods such as Evolutionary Algorithms (EAs) have become the mainstream approach for solving NP-hard optimization problems~\cite{wang2022genetic,zhou2021survey,li2017seeking,beke2024routing}.
Most, if not all, of these methods involve a set of free parameters that would affect their search behavior.
While theoretical analyzes for many search-based methods have offered worst or average bounds on their performance, their actual performance in practice is highly sensitive to the settings of parameters, i.e., algorithm configurations.
However, finding the optimal algorithm configurations requires both domain-specific knowledge and algorithmic expertise, which cannot be done manually with ease.
In response to this, significant efforts have been made to automate the tuning procedure, commonly referred to as automatic parameter tuning~\cite{huang_survey_2020} and automatic algorithm configuration (AAC)~\cite{ansotegui_model-based_2015,lopez-ibanez_irace_2016,
	hutter_sequential_2011}.
Typically, AAC follows a high-level generate-and-evaluate process, where different configurations are iteratively generated and evaluated on a given set of problem instances, i.e., the training set.
Upon termination, the best-performing configuration is returned.
Since an algorithm configuration fully instantiates a parameterized algorithm, for brevity, throughout this article we will use the term ``configuration'' to refer to the resultant algorithm.

Building upon AAC, a series of works have been conducted to identify a set of configurations instead of a single configuration to form a parallel algorithm portfolio (PAP), referred to as automatic construction of PAPs~\cite{lindauer_automatic_2017,liu_automatic_2019,tang_few-shots_2021,liu_generative_2022}.
Each configuration in the PAP is called a member algorithm.
When applied to a problem instance, all member algorithms in the PAP run independently in parallel to obtain multiple solutions, from which the best solution is returned.
Although a PAP consumes more computational resources than a single algorithm, it can achieve superior overall performance than any single algorithm through the complementary strengths of its member algorithms~\cite{huberman1997economics,gomes_algorithm_2001}.
That is, different member algorithms of the PAP excel at solving different types of problem instances.
Moreover, PAPs employ simple parallel solution strategies, making them well-suited to exploit modern computing facilities (e.g., multi-core CPUs).
Nowadays, such capability has become increasingly crucial for tackling computationally challenging problems~\cite{sutton2019bitter}, given the tremendous advancement of parallel computing architectures in the past decade~\cite{hockney2019parallel}.

Generalization is the core objective in the automatic construction of PAPs~\cite{tang_few-shots_2021}.
It requires that the constructed PAP performs well not only on instances within the training set but also on unseen instances from the same problem class.
Given a training set that sufficiently represents the problem class, existing approaches have proven highly effective in constructing PAPs with good generalization~\cite{lindauer_automatic_2017,kadioglu_isac_2010,xu_hydra_2010,liu_automatic_2019}.
However, in real-world applications, one is very likely to encounter the few-shots scenarios, where the collected training instances are limited and biased (e.g., sampled from a specific distribution) and thus fail to sufficiently represent the target problem class~\cite{smith-miles_generating_2015}.
To address this challenge, recent studies have explored integrating instance generation into the construction process~\cite{tang_few-shots_2021,liu_generative_2022,wang_asp_2024}.
One representative approach is the co-evolution of parameterized search (CEPS)~\cite{tang_few-shots_2021} that maintains two competing populations during evolution: a configuration population (PAP) and a problem instance population.
The evolution of the instance population aims to identify challenging instances that exploit weaknesses in the current PAP, while the evolution of the PAP improves its generalization by identifying configurations that better handle these instances.

\begin{figure*}[tbp]
	\centering
	\includegraphics[width=0.8\textwidth]{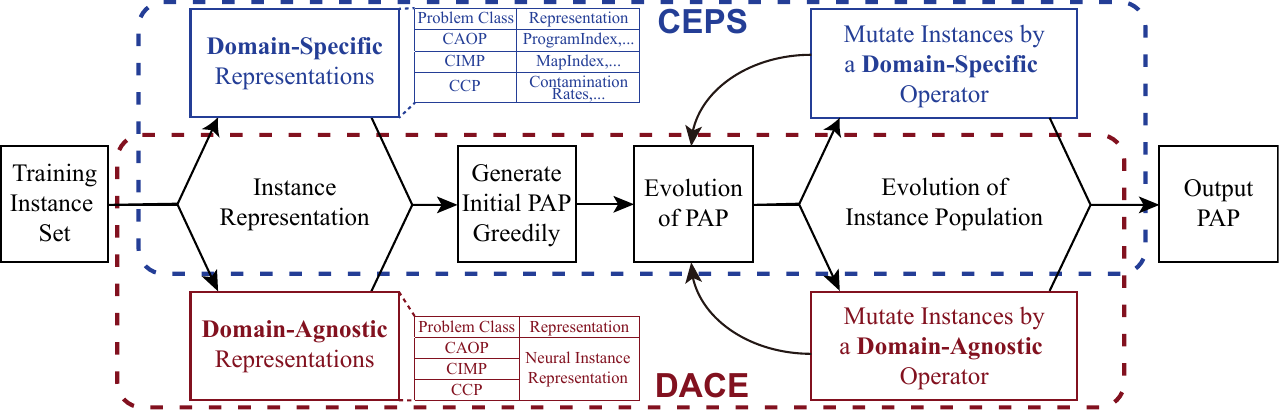}
    \vspace{-5pt}
	\caption{A contrastive view of DACE and CEPS. While both follow the same co-evolutionary framework, CEPS requires domain-specific instance generators when applied to a problem class, whereas DACE employs neural network-based domain-agnostic instance representation and generation.}
	\label{fig:dace_vs_ceps}
    \vspace{-12pt}
\end{figure*}

Although CEPS has shown promising performance in few-shot scenarios, it has one major limitation.
That is, CEPS fundamentally relies on domain-specific instance generators, which significantly limits its generality and practical applicability.
As shown in Fig.~\ref{fig:dace_vs_ceps}, when applying CEPS to a specific problem class, one needs to provide instance generators tailored to the problem class to generate problem instances during the evolution of the instance population.
As demonstrated in~\cite{tang_few-shots_2021}, when applying CEPS to the traveling salesman problem (TSP), Tang~\textit{et al}. developed a TSP-specific instance mutation operator based on 2D coordinate representations of TSP instances.
However, designing such generators requires comprehensive knowledge of the problem class, including suitable instance representations and corresponding variation mechanisms.
This becomes particularly challenging for newly emerging optimization problems and black-box problems where such knowledge is not readily available.
For example, consider the compiler argument optimization problem~\cite{jiang_smartest_2022}, which aims to find the optimal compiler argument settings to minimize the size of the binary file generated from compiling a given program source code.
For such a problem, the instance generator should be able to create new yet valid program source code -- a task that demands deep understanding of program syntax and semantics.

This work proposes a novel PAP construction approach called \textbf{d}omain-\textbf{a}gnostic \textbf{c}o-\textbf{e}volution of parameterized search~(DACE).
The term ``domain-agnostic'' refers specifically to the problem domain.
That is, unlike existing methods requiring domain-specific instance generators  built upon a problem instance's structural information (e.g., the TSP instance generator as described above), DACE has a general-purpose generator that does not need such information.
This allows DACE to be applicable to diverse binary optimization problem classes where variables take 0/1 values.
Such problems are widespread in practice, e.g., appearing in combinatorial optimization\(^1\), discrete facility location, and certain graph problems (e.g., max cut and max coverage).

As shown in Fig.~\ref{fig:dace_vs_ceps}, the key novelty of DACE lies in its universal, domain-agnostic neural network-based representation of problem instances and its mechanism for evolving these neural networks to generate new problem instances.
Additionally, DACE automatically extracts domain-invariant features from training instances and uses these features to constrain instance generation, ensuring the generated instances are meaningful.
In essence, DACE addresses the challenge of representing and generating instances of binary optimization problems without domain-specific knowledge, an area that remains largely unexplored in existing research (as discussed further in Section~\ref{sec:Inst_generation}).

DACE follows the same co-evolutionary framework as CEPS where a configuration population (PAP) and a problem instance population compete with each other; therefore it is particularly well-suited for constructing generalizable PAPs in few-shot scenarios.
Compared to CEPS, DACE offers a significant advantage in terms of generality and practical applicability.
In particular, DACE requires only a small set of training instances from the problem class and no domain-specific instance generators.
Furthermore, and importantly, these training instances can be provided purely as black boxes, where only solution evaluation (i.e., input a solution and output a fitness score) is available without access to the analytic form of the objective function.
This means DACE can readily construct PAPs for black-box optimization problems.
In contrast, existing PAP construction approaches with reliance on domain-specific instance generators face significant challenges with such problems.

The main contributions of this work are summarized below.
\begin{enumerate}
	\item A novel PAP construction approach, namely DACE, is proposed.
	      By eliminating the need for domain-specific instance generators, DACE is a general-purpose, off-the-shelf PAP construction approach for binary optimization problems.
	\item The strong generality of DACE is validated across three real-world binary optimization problems including a black-box problem.
	      Compared to CEPS with expert-crafted domain-specific instance generators, for DACE, a method requiring no domain-specific knowledge, our baseline expectation is to achieve performance comparable to CEPS.
	      Surprisingly, DACE not only matches but actually outperforms CEPS across all three problems.
	\item The proposed domain-agnostic instance representation and generation approach serves as a general data augmentation technique, which not only benefits PAP construction but also opens up new possibilities for training broad classes of optimizers, i.e., learning to optimize (L2O)~\cite{tang2024learn}.
\end{enumerate}

The remainder of this article is organized as follows.
Section~\ref{sec:preliminaries} introduces the problem of seeking generalizable PAPs in few-shot scenarios, as well as existing PAP construction and problem instance generation methods.
Section~\ref{sec:instance_generation} presents DACE's domain-agnostic instance representation and generation, followed by its complete framework in Section~\ref{sec:dace_framework}.
Computational studies are presented in Section~\ref{sec:experiments}.
Section~\ref{sec:conclusion} concludes the article with discussions.

\section{Few-Shot Construction of Generalizable PAPs}
\label{sec:preliminaries}

\subsection{Notations and Problem Description}
Assume a PAP is to be built for a problem class, for which we denote a problem instance as $s$ and the set of all possible instances as \(\Omega\).
Given a parameterized optimization algorithm with its configuration space \(\Theta\), each \(\theta \in \Theta\) is an algorithm configuration that fully instantiates the algorithm.
Let \(P=\left\{\theta_1,\theta_2,\cdots,\theta_K\right\}\) denote a PAP that consists of $K$ configurations as its member algorithms.
For any problem instance \(s \in \Omega\) and configuration \(\theta \in \Theta\), let \(f\left(\theta, s\right)\) denote the performance of \(\theta\) on \(s\).
In practice, this indicator can measure various aspects of performance such as solution quality~\cite{lopez-ibanez_irace_2016}, computational efficiency~\cite{liu_performance_2020}, or even be stated in a multi-objective form~\cite{bolt2016moparamils}.
Without loss of generality, we assume larger values indicate better performance.
The performance of \(P\) on instance \(s\) is the best performance achieved among its member algorithms \(\theta_1,\theta_2,\cdots,\theta_K\) on \(s\):
\begin{equation}
    \label{eq:pap_performance}
    f\left(P, s\right) = \max_{\theta\in P}f\left(\theta, s\right).
\end{equation}
The goal is to identify $K$ configurations $\theta_1,\ldots,\theta_K$ from $\Theta$ to form a PAP $P$ that achieves the optimal generalization performance over $\Omega$:
\begin{align}
    \label{eq:pap_objective_1}
    \max_{P} J\left(P,\Omega\right) = \int_{s\in\Omega}f\left(P,s\right) p(s) \textrm{d}s.
\end{align}
Here, $p(s)$ denotes the prior probability distribution of instances in $\Omega$.
Since the prior distribution is typically unknown in practice, a uniform distribution can be assumed without loss of generality.
Then Eq.~(\ref{eq:pap_objective_1}) simplifies to Eq.~(\ref{eq:pap_objective_2}) by omitting a normalization constant:
\begin{equation}
    \label{eq:pap_objective_2}
    \max_{P} J\left(P,\Omega\right)=\int_{s\in\Omega}f\left(P,s\right) \textrm{d}s.
\end{equation}

In practice, directly optimizing Eq.~(\ref{eq:pap_objective_2}) is intractable since the set \(\Omega\) is generally unavailable.
Instead, only a set of training instances, i.e., a subset $T \subset \Omega$, is given for the construction of $P$.

\subsection{
Existing Approaches for Constructing Generalizable PAPs
}

\label{sec:related_work}
When the training set $T$ is sufficiently large and can effectively represent $\Omega$, one can construct a PAP with good generalization by optimizing its performance on the training set:
\begin{align}
\label{eq:pap_objective_training_set}
\max_{P} J\left(P,T\right) = \sum_{s\in T}f\left(P,s\right).
\end{align}
Existing approaches such as GLOBAL~\cite{lindauer_automatic_2017}, PARHYDRA~\cite{xu_hydra_2010}, CLUSTERING~\cite{kadioglu_isac_2010}, and PCIT~\cite{liu_automatic_2019} have proven effective in such scenarios.
However, when the training instances are rather limited, i.e., few-shot scenarios, optimizing performance solely on the training set can lead to the overtuning phenomenon~\cite{liu_performance_2020}, similar to the overfitting in machine learning.
That is, the constructed PAP performs well on the training set but its test (generalization) performance is arbitrarily bad.

Leveraging synthetic instances during the construction process has been shown to be effective in addressing the above challenge~\cite{tang_few-shots_2021,liu_generative_2022,wang_asp_2024}.
The representative approach is CEPS~\cite{tang_few-shots_2021} that employs a co-evolutionary framework~\cite{ma2018survey} where a configuration population $P$ and an instance population $T$ compete with each other.
Specifically, each round of its co-evolution has two steps:
\begin{enumerate}
\item     Evolution of PAP $P$: with $T$ fixed, identify an improved PAP $P'$ that maximizes $\sum_{s\in T}f\left(P^\prime, s\right)$, and then update $P \leftarrow P'$.
    \item Evolution of training set $T$: with $P$ fixed, generate a new instance set $T'$ that minimizes $\sum_{s\in T^\prime}f\left(P, s\right)$, and then update $T \leftarrow T \cup T'$;

\end{enumerate}
As shown in~\cite{tang_few-shots_2021}, the two-step process effectively maximizes the lower bound, i.e., a tractable surrogate, of PAP's generalization performance as defined in Eq.~(\ref{eq:pap_objective_2}).
While CEPS has shown promising results in few-shot scenarios, applying it to a specific problem class requires one to provide domain-specific instance generators, particularly mutation operators, for evolving instances in the first step.

\subsection{Existing Problem Instance Generation Methods}
\label{sec:Inst_generation}
Designing effective instance generators is crucial for the few-shot construction of generalizable PAPs, yet it is often difficult, as discussed earlier.
Domain-specific generators developed for particular problems, such as 
TSP~\cite{tang_few-shots_2021,Hemert06}, vehicle routing problem~\cite{tang_few-shots_2021}, and job shop scheduling problem~\cite{branke2011evolutionary}, can be utilized but are limited only to those specific problem domains.
In deep learning, instance generation methods have also been developed to address few-shot scenarios in vision ~\cite{zhang2022fewshot} and natural language domains~\cite{zhou2022fipda}.
However, due to the fundamental differences between these domains (images and text) and optimization problems, these approaches cannot be applied to the latter. 
Furthermore, for emerging and black-box optimization problems where even experts lack sufficient domain knowledge, designing domain-specific instance generators becomes even more challenging.
The following sections will present DACE that eliminates the need for domain-specific instance generators.

\section{Domain-Agnostic Problem Instance Representation and Generation}
\label{sec:instance_generation}
The key innovation of DACE lies in its domain-agnostic approach to representing and generating problem instances.
At its core, DACE employs neural networks (NNs) as a universal representation for problem instances, dubbed neural instance representation (NIR).
Given a small set $T$ of training instances, DACE first converts these instances into NIRs and then uses these NIRs as a basis for generating new instances, which are also represented as NIRs.
Since different problem instances essentially differ in their underlying objective functions, specifically, in how they map solutions to objective values.
Therefore, an NIR represents a problem instance by approximating its objective function.
This way, the training process of an NIR only requires pairs of solutions and their corresponding objective values, without requiring any domain-specific knowledge such as analytic form of the objective function.
Furthermore, by varying the weight parameters within the NIR, different objective functions corresponding to different problem instances are obtained.

As a result, NIR maintains generality across various binary optimization problem classes, where training instances in \(T\) can be treated as black boxes that simply return solution evaluations.
It is worth noting that the term ``generality'' differs from ``generalization''.
Specifically, PAP's generalization indicates its ability to perform well not only on training instances, but also on unseen instances from the same problem class, whereas NIR's generality means its broad applicability to various binary optimization problem classes.

Technically, NIRs share similarities with surrogate models, a technique widely used in EAs to reduce the number of expensive fitness evaluations~\cite{Jin11,zhang2021surrogate,nguyen2024constrained,lin2022ensemble}.
However, it is important to note that NIRs serve a fundamentally different purpose here.
Rather than reducing number of fitness evaluations, NIRs serve as the basis for instance generation, where all generated instances in DACE are represented as NIRs instead of their actual forms.
To achieve this goal, two technical challenges must be addressed.
First, effective mutation operators need to be developed that can generate meaningful instances represented as NIRs.
Due to the powerful representation capabilities of NNs, particularly deep NNs, random mutations of NIR parameters could produce arbitrary objective functions that do not correspond to valid actual problem instances.
In other words, the generated instances should relate to the training instances and belong to the same problem class.
This requires the extraction and utilization of domain-invariant features from the training instances to properly constrain the instance generation process.
The second challenge arises from the discrete nature of binary optimization problems.
In these problems, small changes in discrete inputs (such as flipping a few bits) can result in dramatic changes in objective values.
This makes NN learning particularly challenging, as NNs are typically suited for fitting smooth, continuous functions~\cite{ferrari2005smooth}.

To address these challenges, we propose a decoupled design of the structure of NIR based on variational autoencoders (VAE)~\cite{kingma_auto-encoding_2014} and hypernetworks~\cite{chauhan_brief_2024}.
This structure decouples domain-invariant features from instance-specific features, and also decouples solution encoding from objective function approximation.
Below we detail the NIR structure, its training method, and the NIR-based instance mutation operator.

\subsection{Structure of the NIR}
\label{subsec:surrogate model structure}

\begin{figure}[tbp]
    \centering
    \includegraphics[width=0.8\linewidth]{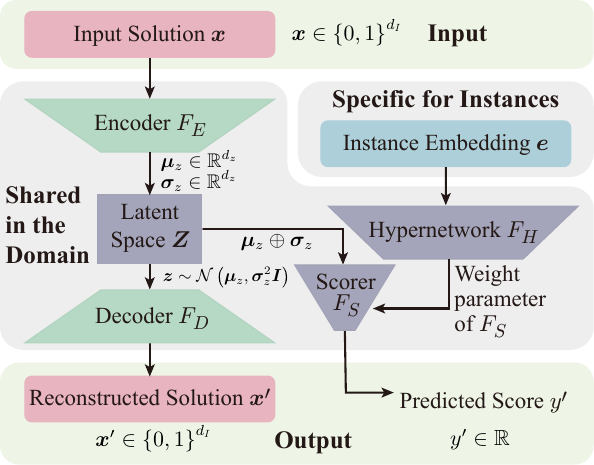}
    \caption{An overview of the NIR for a problem instance. }
    \label{fig:surrogate}
    \vspace{-12pt}
\end{figure}

As shown in Fig.~\ref{fig:surrogate}, NIR employs a VAE for encoding discrete solutions into continuous latent vectors and decoding them back.
The VAE consists of an encoder \(F_E\) and a decoder \(F_D\), which are both multilayer perceptrons (MLPs).
A scorer \(F_S\), also implemented as an MLP, is built upon the latent vectors output by \(F_E\) to approximate the objective function of the problem instance.
Using real-valued latent vectors instead of original discrete solutions as inputs to \(F_S\) creates a smoother function mapping that NNs can approximate more effectively.
Let \(\boldsymbol{w}_E\) and \(\boldsymbol{w}_D\) denote the weight parameters of \(F_E\) and $F_D$, respectively.
Given an input solution \(\boldsymbol{x}\in\left\{0, 1\right\}^{d_I}\), where \(d_I\) is the dimension of the problem instance,  \(F_E\) predicts the means \(\boldsymbol{\mu}_z\in\mathbb{R}^{d_z}\) and standard deviations \(\boldsymbol{\sigma}_z\in\mathbb{R}^{d_z}\) of a \(d_z\)-dimensional multivariate Gaussian distribution \(\mathcal{N}\left(\boldsymbol{\mu}_z,\boldsymbol{\sigma}_z^2\boldsymbol{I}\right)\).
A vector $\boldsymbol{z} \in \mathbb{R}^{d_z}$ is then sampled from the distribution:
\begin{equation}
\label{eq:encoder}
\begin{aligned}
\left[\boldsymbol{\mu}_z,\boldsymbol{\sigma}_z\right]=F_E\left(\boldsymbol{w}_E; \boldsymbol{x}\right)\\ \boldsymbol{z}\sim\mathcal{N}\left(\boldsymbol{\mu}_z,\boldsymbol{\sigma}_z^2\boldsymbol{I}\right).
\end{aligned}
\end{equation}
Based on $\boldsymbol{z}$, the decoder $F_D$ outputs a reconstructed solution $\boldsymbol{x}^\prime \in \left\{0, 1\right\}^{d_I}$:
\begin{equation}
\label{eq:decoder}
\begin{aligned}
\boldsymbol{x}^\prime = F_D(\boldsymbol{w}_E; \boldsymbol{z}).
\end{aligned}
\end{equation}
Let \(\boldsymbol{w}_S\) denote the weight parameters of \(F_S\), the concatenation (denoted by \(\oplus)\) of \(\boldsymbol{\mu}_z\) and \(\boldsymbol{\sigma}_z\) is fed into \(F_S\), which predicts the score (objective value) \(y^\prime\):
\begin{equation}
\label{eq:scorer}
\begin{aligned}
 y^\prime=F_S\left(\boldsymbol{w}_S; \boldsymbol{\mu}_z\oplus\boldsymbol{\sigma}_z\right).        
\end{aligned}
\end{equation}

To capture domain-invariant features of the problem class, the encoder $F_E$ and decoder $F_D$ are shared across all NIRs, i.e., all instances in the class.
Additionally, the scorer's parameters $\boldsymbol{w}_S$ are generated by a hypernetwork $F_H$, which is also shared across all NIRs.
Specifically, $F_H$ is an MLP with weight parameters \(\boldsymbol{w}_H\) that takes an instance-specific embedding \(\boldsymbol{e} \in \mathbb{R}^{d_e}\) as input and outputs $\boldsymbol{w}_S$:
\begin{equation}
\boldsymbol{w}_S = F_H(\boldsymbol{w}_H;\boldsymbol{e}).
\end{equation}
This means the weight parameters \(\boldsymbol{w}_S\) of the scorer \(F_S\) is generated by the hypernetwork \(F_H\) through its output conditioned on the input embedding \(\boldsymbol{e}\).
The instance embedding \(\boldsymbol{e}\) is a trainable vector optimized to encode instance-specific information.
It is randomly initialized and then trained with (solution, objective value) pairs sampled from the corresponding training problem instance, as detailed in Section~III-B.

In summary, the trainable parameters in an NIR include \(\boldsymbol{w}_E\), \(\boldsymbol{w}_D\), \(\boldsymbol{w}_H\), and the instance embedding \(\boldsymbol{e}\).
The first three are shared across all NIRs, while the last one is instance-specific.
By training these parameters on the training instances, domain-invariant features are automatically extracted and encoded into \(\boldsymbol{w}_E\), \(\boldsymbol{w}_D\), and \(\boldsymbol{w}_H\), while instance-specific features are captured in their respective embeddings.
When generating new instances represented as NIRs, the shared parameters are kept fixed while only instance embedding is perturbed, ensuring that the learned domain-invariant features are preserved.

\begin{algorithm}[tbp]
    \small
    \caption{NIR-based Instance Mutation Operator}
    \label{alg:instance mutation}
    \SetKwInput{KwData}{Input}
    \SetKwInput{KwResult}{Output}
    \KwData{Problem instance represented as NIR \(m\), PAP \(P\), black-box continuous optimizer $OPT$.}
    \KwResult{Mutated instance represented as NIR \(m^\prime\).}
    \SetKw{To}{to}
    \SetKw{Append}{append}
    \SetKw{Into}{into}
    \SetKw{Break}{break}
    \SetKw{Return}{return}
    \SetKwProg{Fn}{Function}{:}{end}
    \SetKwFunction{Sort}{sort}
    \SetKwFunction{Normalize}{normalize}
    \SetKwFunction{Distance}{distance}
    \SetKwFunction{MSE}{MSE}
    \SetKwFunction{MLP}{MLP}
    \SetKwFunction{InsFeature}{ins\_feature}
    \SetKwComment{EmptyLine}{ }{ }
    \SetNoFillComment
    Initialize \(OPT\) with instance embedding \(\boldsymbol{e}\) of \(m\)\;
    \(\boldsymbol{e}^{\prime} \leftarrow \) minimize Eq.~(\ref{eq:instance_generation}) using \(OPT\);\\
    \lIf{\(f\left(P, m | \boldsymbol{e}\right) \leq f\left(P,m^\prime | \boldsymbol{e}^\prime\right)\)}
    {\(\boldsymbol{e}^\prime \leftarrow \boldsymbol{e}\)}
    \Return \(m^\prime\textrm{ specified by }\boldsymbol{e}^\prime\)\;
\end{algorithm}

\subsection{Training NIRs}
\label{subsec:surrogate model training}
Given a small set of training instances \(T = \{s_1, s_2, \ldots\}\), an NIR is built for each \(s_i\), denoted as \(m_i\).
As described earlier, all NIRs share \(\boldsymbol{w}_E\), \(\boldsymbol{w}_D\), and \(\boldsymbol{w}_H\), but each has its own instance embedding.
Let $\boldsymbol{e}_i$ denote the embedding of $s_i$.
For each instance \(s_i\), a set of solutions and their corresponding objective values is assumed to be available, denoted as \(\mathcal{X}^i = \{(\boldsymbol{x}_1^i, y_1^i), (\boldsymbol{x}_2^i, y_2^i), \dots\}\).
This set can be obtained by running search or sampling methods on $s_i$.
Given \(\mathcal{X}^i\) \((i=1,2,\ldots, |T|)\), all parameters \(\boldsymbol{w}_E\), \(\boldsymbol{w}_D\), \(\boldsymbol{w}_H\), and \(\boldsymbol{e}_i\ (i=1,2,\ldots, |T|)\) are trained by minimizing the following loss, where $\lambda_1$ and $\lambda_2$ are weighting hyper-parameters and MSE denotes the mean squared error:
\begin{equation}
\label{eq:loss}
    \begin{aligned}        \min_{\substack{\boldsymbol{w}_E,\boldsymbol{w}_D,\boldsymbol{w}_H,\\\boldsymbol{e}_1,\boldsymbol{e}_2,\cdots,\boldsymbol{e}_{\left\vert T\right\vert}}}
          \sum_{i=1}^{\left\vert T\right\vert}
        \sum_{(\boldsymbol{x}, y) \in \mathcal{X}^i}
        &\textrm{MSE}\left(\boldsymbol{x}, \boldsymbol{x}^\prime\right)+\lambda_1\textrm{MSE}\left(y, y^\prime\right) \\
        +\lambda_2&\mathbb{D}_\textrm{KL}\left(\mathcal{N}\left(\boldsymbol{\mu}_z,\boldsymbol{\sigma}_z^2\boldsymbol{I}\right), \mathcal{N}\left(\boldsymbol{0},\boldsymbol{I}\right)\right).\\
    \end{aligned}
\end{equation}
Here, \(\boldsymbol{\mu}_z,\boldsymbol{\sigma}_z\) are obtained through Eq.~(\ref{eq:encoder}), and \(\boldsymbol{x}^\prime, y^\prime\) are obtained through Eq.~(\ref{eq:decoder}) and Eq.~(\ref{eq:scorer}), respectively.
The loss function consists of three terms.
The first term is a reconstruction loss between the input solution \(\boldsymbol{x}\) and its reconstructed counterpart \(\boldsymbol{x}^\prime\), promoting the encoder to capture the structure of \(\boldsymbol{x}\).
The second term is a prediction loss for objective values, encouraging the scorer to be accurate.
It should be noted that the actual objective value \(y\) has been normalized to the range \(\left[0,1\right]\) to prevent the influence of varying value scales across different problem instances on the training process.
The third term is the KL-divergence between the learned latent distribution and the standard Gaussian distribution, which serves as a standard regularization term commonly used in VAEs~\cite{kingma_auto-encoding_2014} to ensure a smooth latent space. 
All parameters are randomly initialized and jointly optimized using stochastic gradient descent.

During the training process of NIR, constraint information from problem instances is not incorporated, meaning NIR does not directly handle constraints.
This is because constraints in optimization problems are typically problem-dependent, and encoding them into the model's representation would compromise the generality of NIR.
To address constrained optimization problems,  constraint-handling techniques are applied to convert infeasible solutions into feasible ones before they are fed into NIR.
More details can be found in~Section~\ref{subsec: domain and instance generation}.

\begin{algorithm}[tbp]
    \small
    \caption{DACE}
    \label{alg:dace}
    \SetKwInput{KwData}{Input}
    \SetKwInput{KwResult}{Output}
    \KwData{Training set \(T\); number of member algorithms \(K\); maximum number of configuration mining iterations $n$; maximum round number \(MaxRound\).}
    \KwResult{The final configuration population \(P\)}
    \SetKw{To}{to}
    \SetKw{Append}{append}
    \SetKw{Into}{into}
    \SetKw{Break}{break}
    \SetKwProg{Fn}{Function}{:}{end}
    \SetKwFunction{Sort}{sort}
    \SetKwFunction{Normalize}{normalize}
    \SetKwFunction{Distance}{distance}
    \SetKwFunction{MSE}{MSE}
    \SetKwFunction{MLP}{MLP}
    \SetKwFunction{Len}{len}
    \SetKwFunction{InsFeature}{ins\_feature}
    \SetNoFillComment
    \tcc{--------Initialization--------}
    \(M\leftarrow\) build an NIR for each problem instance in \(T\);\\
    Randomly sample a subset \(C \subset \Theta\) and test all selected configurations on each NIR in \(M\);\\
    \(P\leftarrow \emptyset\);\\
    \For{\(i\leftarrow 1\) \To \(K\)}{
        Find \(\theta_i \in C\) that maximizes \(
        \sum\nolimits_{m\in M}f\left(P\cup\left\{\theta_i\right\}, m\right)\)\;
        \(P\leftarrow P\cup\theta_i\)
    }

    \tcc{----------Co-Evolution----------}
    \For{\(r \leftarrow\) 1 \To \(MaxRound\)}{
        \tcc{----Evolution of \(P\)----}
        \(\Psi \leftarrow P\)\;
        \For{\(i\leftarrow 1\) \To \(n\)}{
            \(j\leftarrow i \mod K\)\;
            \(P^\prime\leftarrow P \setminus \{\theta_j\}\)\;
            Use an AAC procedure to identify \(\theta^{\prime}\in \Theta\) that maximizes $\sum\nolimits_{m\in M} f\left(P^\prime\cup\left\{\theta^\prime\right\}, m\right)$\;
            \(\Psi \leftarrow \Psi \cup\left\{\theta^\prime\right\}\)\;
        }
        Identify $\theta_1,...,\theta_K $ from $\Psi$ to form $P$ that maximizes
        \(\sum\nolimits_{m\in M}f\left(P, m\right) \);\\
       \tcc{----Evolution of \(M\)----}
        \lIf{\(r=MaxRound\)}{\Break}
        Assign the fitness of each $m \in M$ as $-f(P, m)$;\\
        \(M^{\prime}\leftarrow\) create a copy of \(M\)\;
        \(M_{new}\leftarrow\emptyset\)\;
        \For{\(i \leftarrow 1\) \To\(|M^\prime|/2\)}{
            \(m^\prime\leftarrow\)  Randomly selected \(m\in M^\prime\) and \textbf{mutate} $m$ with Alg.~\ref{alg:instance mutation};\\
            Test $P$ on $m^\prime$ and assign the fitness of $m^\prime$ as $-f(P, m^\prime)$;\\
            \(m^\star\leftarrow\) randomly select one from all the NIRs in \(M\) with lower fitness than \(m^\prime\)\;
            \lIf{\(m^\star\) not found}{\Break}
            \(M\leftarrow M \backslash \left\{m^\star\right\}\)\;
            \(M_{new}\leftarrow M_{new}\cup \left\{m^\prime\right\}\)\;
        }
        \(M\leftarrow M_{new} \cup M^\prime\)\;
    }
    \Return{\(P\)}
\end{algorithm}

\subsection{NIR-based Instance Generation and Evaluation}
\label{subsec:instance_generation}
Alg.~\ref{alg:instance mutation} presents the NIR-based instance mutation operator, the key mechanism for generating new instances (NIRs) during the evolution of DACE's instance population.
This operator takes an NIR $m$ as input and outputs a new NIR $m^\prime$ that is challenging for the PAP $P$.
Specifically, $m^\prime$ is generated by perturbing the instance embedding $\boldsymbol{e}$ of $m$ to find a new embedding $\boldsymbol{e}^\prime$ that minimizes Eq.~(\ref{eq:instance_generation}):
\begin{equation}
\label{eq:instance_generation}
\min_{\boldsymbol{e}^\prime \in \mathbb{R}^{d_e}} f(P, m^\prime | \boldsymbol{e}^\prime),
\end{equation}
where $f(P, m^\prime | \boldsymbol{e}^\prime)$ is the indicator $f$ measuring the performance of PAP $P$ on the NIR $m^\prime$ specified by $\boldsymbol{e}^\prime$, which is the best performance achieved among $P$'s member algorithms on $m^\prime$, as defined in Eq.~(\ref{eq:pap_performance}). A smaller value indicates that the NIR is more challenging for $P$.

In summary, our objective is to find a new embedding \(\boldsymbol{e}^\prime\) such that the NIR \(m^\prime\) with this embedding becomes more challenging for the PAP \(P\).
It is noted that during the evolutionary process of DACE, we directly employ the NIR \(m^\prime\) to evaluate the performance of algorithm configurations rather than reverting the NIR \(m^\prime\) back to an authentic problem instance in the problem class \(\Omega\).
In fact, such a restoration cannot be achieved without domain-specific knowledge.

When the performance indicator $f$ concerns solution quality, it is important to normalize the objective values of solutions found by the PAP to make $f$ values comparable across different NIRs.
To achieve this normalization, we leverage the computational efficiency of NNs on massive-parallel GPUs to randomly sample a large number (10M) solutions and evaluate their objective values using the NIR $m^\prime$.
The maximum and minimum objective values (denoted as $f^{max}_{{m}^\prime}$ and $f^{min}_{{m}^\prime}$, respectively) from these samples are used to normalize the original objective value $f^{ori}_{\boldsymbol{m}^\prime}$ obtained by the PAP on $m^\prime$:
\begin{equation}
\label{eq:normalize}
f(P, m^\prime| {\boldsymbol{e}^\prime}) = \frac{f^{ori}_{m^\prime} - f^{min}_{m^\prime}}{f^{max}_{m^\prime} - f^{min}_{m^\prime}}.
\end{equation}

Since the performance indicator $f$ typically lacks analytic forms and $\boldsymbol{e}^\prime$ is a real-valued vector, Eq.~(\ref{eq:instance_generation}) represents a continuous black-box optimization problem.
In this work, PGPE~\cite{beyer_evolution_2002}, an evolution strategy (ES) method, is employed to optimize Eq.~(\ref{eq:instance_generation}) (lines 1-2 in Alg.~\ref{alg:instance mutation}).
Details of PGPE are provided in Appendix~A of the supplementary.
Note that other black-box continuous optimizers could also be used here, as the specific choice of optimizer is not central to our approach.
Upon termination, the mutation operator returns the NIR $m^\prime$ specified by the best embedding $\boldsymbol{e}^\prime$ found by the optimizer (lines 3-4 in Alg.~\ref{alg:instance mutation}).

\section{Domain-Agnostic Co-Evolution of Parameterized Search (DACE)}
\label{sec:dace_framework}

\begin{figure*}[tbp]
    \centering
    \includegraphics[width=0.75\textwidth]{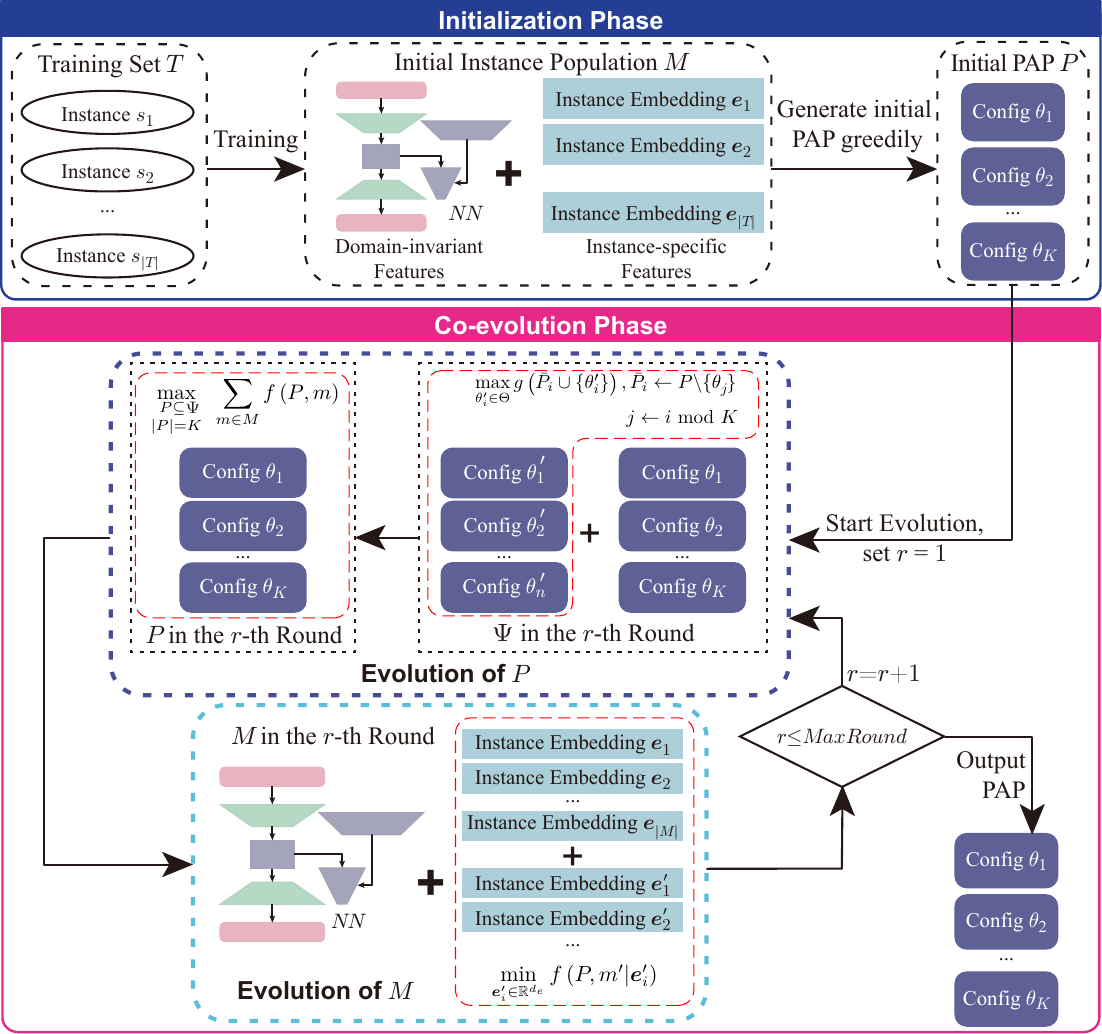}
    \caption{An overview of DACE. It consists of an initialization phase followed by a co-evolution phase where the configuration population $P$ and the instance population $M$ evolve alternately.}
    \label{fig:dace}
    \vspace{-12pt}
\end{figure*}

DACE is a general-purpose, off-the-shelf approach for constructing PAPs for binary optimization problems.
Same as CEPS~\cite{tang_few-shots_2021}, DACE evolves two competing populations: a configuration population (PAP) and a problem instance population.
However, in DACE, problem instances are represented as NIRs, and are generated through the mutation operator described in Section~\ref{sec:instance_generation}.
Additionally, configurations in the PAP are identified from a configuration space defined by a parameterized algorithm.
To eliminate the need for domain-specific parameterized algorithms, general-purpose EAs are used in DACE.
Specifically, the biased random-key genetic
algorithm (BRKGA)~\cite{goncalves_biased_2011} is employed here, which has five configurable parameters: elite population size, offspring population size, mutant population size, elite bias for parent selection, and duplicate elimination flag.
Details of these parameters are provided in Appendix~B of the supplementary.
The configuration space $\Theta$ consists of all possible combinations of these parameter values.

The pseudocode of DACE is presented in Alg.~\ref{alg:dace}, with its workflow diagram illustrated in Fig.~\ref{fig:dace} for better understanding.
Overall, DACE consists of two phases: an initialization phase  followed by a co-evolution phase where PAP and instance population evolve alternately for $MaxRound$ rounds.
These phases are detailed below.

\subsubsection{Initialization Phase (lines 1-7 in Alg.~\ref{alg:dace})}
Given a training set \(T\), an NIR is built for each instance in \(T\) as described in Section~\ref{subsec:surrogate model training}, and the population of NIRs (instances) is denoted as \(M\) (line 1).
To initialize a PAP $P$ of $K$ member algorithms, a set of configurations $C$ is first randomly sampled from BRKGA's configuration space $\Theta$, and these configurations are tested on each NIR in \(M\) (line 2).
Then, the PAP is constructed by iteratively selecting one configuration from $C$ that provides the largest performance improvement on $M$ in terms of the performance indicator $f$, until $K$ configurations are chosen (lines 3-7).

\subsubsection{Evolution of PAP (lines 9-16 in Alg.~\ref{alg:dace})}
\label{subsubsec: EVO_PAP}
Given the current PAP \(P\), a so-called configuration mining process is iterated \(n\) times (lines 10-15).
In the \(i\)-th iteration, the \(j\)-th member algorithm is removed from $P$ to form \(P'\), where \(j=i\mod K\) (lines 11-12).
Then, an AAC procedure (SMAC~\cite{hutter_sequential_2011} is used here, following CEPS) is employed to search for a new configuration \(\theta'\) within \(\Theta\) that maximizes $\sum\nolimits_{m \in M}f(P'\cup \{\theta'\},m)$ (line 13), i.e., the performance of \(P'\cup\left\{\theta'\right\}\) on the instance population \(M\).
After $n$ iterations, $n$ new configurations are obtained.
In CEPS, the configuration whose inclusion yields the best-performing PAP is selected from these $n$ configurations to update $P$, and it is analogous to a mutation operation where the AAC procedure acts as a mutation operator that perturbs one member algorithm in $P$.
DACE extends this approach.
Specifically, a configuration set $\Psi$ is constructed by combining all $n$ new configurations with the $K$ configurations in $P$ (line 9 and line 14).
DACE then examines all possible combinations of $K$ configurations from $\Psi$  and selects the combination with the best performance on $M$ to replace the current PAP (line 16). 
Since the performance of each configuration in $\Psi$ on each NIR in $M$ has already been evaluated, this enumeration does not require actually running the configurations on NIRs, but simply queries the previously recorded performance results.
Given that \(\left\vert\Psi\right\vert=K+n\), the total number of combinations to examine is a combinatorial value\(\binom{n+K}{K}=\frac{\left(n+K\right)!}{K!n!}\), and this would take negligible time when $K$ and $n$ are not large (in our experiments, $K = 4$ and $n = 20$, with \(\binom{n+K}{K}=10626\)).
As a result, this approach can potentially update all configurations in $P$ simultaneously, which is a more powerful mutation operation compared to CEPS's single-configuration mutation.
Since this improvement is not the primary focus of this work, the same PAP mutation mechanism is also applied to CEPS in our experiments to enhance its performance.

\subsubsection{Evolution of Instance Population (lines 17-28 in Alg.~\ref{alg:dace})}
The objective of evolving $M$ is to identify new NIRs that are challenging for the current PAP \(P\).
The fitness of each NIR $m$ is defined as $-f(P,m)$ (line 18).
That is, NIRs on which $P$ performs poorly have higher fitness values.
DACE begins by creating a copy $M'$ of the instance population $M$ (line 19).
Additionally, an empty set \(M_{new}\) is initialized to store newly generated NIRs (line 20).
DACE then repeats a so-called instance mining process \(\left\vert M^\prime\right\vert/2\) times (lines 22-27).
In each iteration, the mutation operator described in Alg.~\ref{alg:instance mutation} is applied to an NIR \(m\) randomly selected from \(M^\prime\) to generate a new NIR \(m^\prime\) (line 22).
If no NIR in \(M\) has lower quality than \(m^\prime\), indicating that the generated instance is not challenging for the current PAP, the mining process terminates (lines 24–25).
Otherwise, an instance \(m^\star\) with lower fitness than \(m^\prime\) is randomly removed from $M$, and \(m^\prime\) is added to \(M_{new}\) (lines 26-27 ).
When the mining process completes, the newly generated NIRs in \(M_{new}\) are added to the instance population \(M\) (line 29), which is used to obtain the new PAP in the next co-evolution round.
The evolution of $M$ will be skipped in the last co-evolution round (line 17) because there is no need to generate more NIRs since the final PAP has been constructed completely.

\section{Computational Studies}
\label{sec:experiments}

Extensive experiments are conducted on three binary optimization problems from real-world applications: the complementary influence maximization problem (CIMP)~\cite{lu_competition_2015}, compiler arguments optimization problem (CAOP)~\cite{jiang_smartest_2022}, and contamination control problem (CCP)~\cite{hu_contamination_2010}.
Through the experiments, we aim to assess the potential of DACE by addressing the following research questions (RQs):
\begin{enumerate}
\item \textit{RQ1}: How does DACE perform across different problem classes, particularly in its ability to match the performance of CEPS that uses domain-specific instance generators?
\item \textit{RQ2}: How do the PAPs constructed by DACE, which consist of general-purpose search methods as member algorithms, perform against state-of-the-art domain-specific optimizers?
\item \textit{RQ3}: How effectively do the generated NIRs represent actual instances from the problem class?
\end{enumerate}
For each problem class, problem instances were generated based on public benchmarks and divided into training and test sets.
Following the experimental setup of CEPS~\cite{tang_few-shots_2021}, few-shot scenarios were simulated where the training set size is limited.
Specifically, the training set for each problem class contained 5 instances, while the test set contained 100 instances with problem dimensions equal to or larger than training instances.
Throughout the experiments, test instances were used solely to evaluate the generalization performance of PAPs obtained by DACE and the compared methods.
The training instances were only used for PAP construction, regardless of which method used.
The source code and benchmark instances for each problem class have been anonymously open-sourced at \url{https://github.com/MetaronWang/DACE}.

It is important to note that for each of the three problem classes, CEPS's instance mutation operator (i.e., its instance generator) directly employs the generation mechanism of the training instances (see details in the next subsection), producing in-distribution problem instances identical to the training set.
This results in an effective mutation operator that assumes CEPS's users have comprehensive domain-specific knowledge of these three problem classes.
Therefore, for DACE -- a method requiring no domain-specific knowledge -- the baseline expectation was to match the performance of CEPS in both \textit{RQ1} and \textit{RQ3}.
Surprisingly, experimental results demonstrate that DACE even outperforms CEPS across all three problems in \textit{RQ1} while achieving comparable performance in \textit{RQ3}.
The problem classes, compared methods, experimental protocol, and results are detailed in the following.

\subsection{Constraint Handling, Problem Classes and Instances}
\label{subsec: domain and instance generation}

During the three problems, CIMP and CCP involve constraints.
As noted in Section~\ref{subsec:surrogate model training}, NIR does not handle constraints directly.
Instead, problem-specific constraint-handling techniques are applied here.
Specifically, solution repair~\cite{coello2002theoretical} is employed to handle the constraint on the size of the selected seed set in CIMP.
For CCP, following the previous study~\cite{oh_combinatorial_2019},
the contamination probability constraint is converted to a penalty component.
Details on each problem class, including their objective functions, constraint-handling techniques, and problem instances, are provided below.

\subsubsection{Complementary Influence Maximization Problem (CIMP)}
The rapid growth of online social platforms has made influence maximization an increasingly important problem.
Given a social network \(\mathcal{G} = \left(V, E, p\right)\), where \(V\) denotes the nodes, \(E\) denotes the edges, and \(p\) specifies influence probabilities between nodes, the influence maximization problem (IMP) aims to find a seed set $S$ of $k$ nodes (each selected node is called a ``seed'') to maximize the expected number of active nodes.
CIMP~\cite{lu_competition_2015} extends IMP by introducing complementary users, making the influence analysis more complex due to collaborator interactions.
Specifically, given a social network \(\mathcal{G} \) and a complementary seed set $S_A$ for opinion $A$, CIMP aims to find a seed set \(S_B\) of \(k\) nodes from a candidate set \(C \subset V\) with \(\left\vert C\right\vert = d_I\) to maximize the spread of opinion \(B\), making it a $d_I$-dimensional binary optimization problem.
The interaction between opinions $A$ and $B$ is governed by parameters \(q_{A|\emptyset}\), \(q_{A|B}\), \(q_{B|\emptyset}\), \(q_{B|A}\), as detailed in \cite{lu_competition_2015}.

In the binary solution representation  \(\boldsymbol{x} \in \{0,1\}^{d_I}\), \(x_i = 1\)  indicates the \(i\)-th node is included into \(S_B\), while \(x_i = 0\) indicates it is excluded.
The objective function is to maximize the expected number of nodes activated by \(S_B\), subject to selecting at most $k$ seeds:
\begin{equation}
\label{eq:CIMP}
    \begin{gathered}
        \max_{\boldsymbol{x}\in\left\{0,1\right\}^{d_I}} \textrm{ActiveNum}\left(\mathcal{G}, S_A, C, q_{A|\emptyset}, q_{A|B}, q_{B|\emptyset}, q_{B|A}, \boldsymbol{x}\right)\hspace{-10pt} \\
        s.t.\quad                                      \Sigma_{i=1}^{d_I}x_i  \leq k 
    \end{gathered}
\end{equation}
Given a solution, if it exceeds \(k\) selected seeds, it is repaired by keeping only the first \(k\) chosen seeds and discarding the rest.
Then, Eq.~(\ref{eq:CIMP}) is evaluated through Monte Carlo simulations of the propagation process on the network.

For CIMP, problem instances were generated using three public social network benchmarks: Wiki~\cite{leskovec2010predicting}, Facebook~\cite{mcauley2012learning}, and Epinions~\cite{leskovec2010predicting}.
The Wiki benchmark was used for generating training instances, while test instances were generated from Epinions, Facebook and Wiki benchmarks, with the benchmark being randomly selected for each instance.
For each \(d_I\)-dimensional CIMP instance, a candidate seed set \(C\) of size \(d_I\) was randomly selected from the node set \(V\).
The interaction parameters \(\left\{q_{A|\emptyset}, q_{A|B}, q_{B|\emptyset}, q_{B|A}\right\}\) were randomly selected from two default settings: \(\left\{0.5, 0.75, 0.5, 0.75\right\}\), and \(\left\{0.5, 0.25, 0.5, 0.25\right\}\). The maximum seed set size \(k\) is an integer randomly selected from \(\left[0.2d_I, 0.6d_I\right]\).
Finally, the training set contained five problem instances with $d_I = 80$, and the test set contained 100 instances evenly split between $d_I=80$ and $d_I=100$.

\subsubsection{Compiler Arguments Optimization Problem (CAOP)}

CAOP~\cite{jiang_smartest_2022} aims to minimize executable file size during software compilation, which is crucial in storage-limited environments.
Software developers use compiler arguments to control various features that can reduce code size.
The impact of these arguments varies depending on the source code being compiled.
While appropriate arguments can effectively reduce file size, poor choices may lead to increased size. Therefore, selecting the right compiler arguments is critical for optimal results.

In a \(d_I\)-dimensional CAOP instance, \(d_I\) compiler arguments are considered, and the source code to be compiled is denoted as \(F\).
The solution \(\boldsymbol{x} \in \{0,1\}^{d_I}\) represents which arguments are used.
If \(x_i = 1\), the \(i\)-th compiler argument is enabled, whereas \(x_i = 0\) indicates it is disabled. 
The objective function is defined as the negative of the generated executable file size (the negative sign is used to convert the minimization problem into a maximization problem):
\begin{equation}
\label{eq:caop}
    \max_{\boldsymbol{x} \in\left\{0,1\right\}^{d_I}}  -\text{ExeSize}\left(F, \boldsymbol{x}\right).
\end{equation}
Due to the complex nature of the compilation process, Eq.~(\ref{eq:caop}) lacks an analytic form and can only be evaluated by actually compiling the source code $F$  with the specified arguments, making CAOP a black-box optimization problem.

For CAOP, problem instances were generated using the cbench and polybench-cpu~\cite{fursin2009collective} benchmarks, which contain 50 program source files in total.
Among them 11 files were selected as training group and 26 files were selected as test group.
To generate a training instance, a source file was randomly selected from the training group, while test instances were generated using random selections from the test group.
Following~\cite{jiang_smartest_2022}, GCC was chosen as the compiler due to its widespread use.
Among GCC's 186 available arguments, \(d_I\) arguments were randomly selected for each instance, with all other arguments being disabled.
Finally, the training set contained five problem instances with \(d_I=80\), and the test set contained 100 instances in total, evenly split between \(d_I=80\) and  \(d_I=100\).

\subsubsection{Contamination Control Problem (CCP)}
CCP~\cite{hu_contamination_2010,oh_combinatorial_2019} arises from the need for contamination prevention in the food production supply chain.
Multiple stages are involved during food production, each of which can potentially introduce contamination.
At stage $i$, taking mitigation measures can reduce contamination at a rate of random variable \(\Gamma_i\), but incurs a cost of \(c_i\). 
If no action is taken, the contamination rate will be \(\alpha_i\).
In a $d_I$-dimensional CCP instance, there are $d_I$ stages.
For a binary solution  \(\boldsymbol{x} \in \{0,1\}^{d_I}\), \(x_i = 1\)  indicates measures are taken at stage $i$ while $x_i=0$ representing no measures.
Let \(z_i\) denote the proportion of contaminated food at stage \(i\), which is defined as \(z_i = \alpha_i (1 - x_i)(1 - z_{i-1}) + (1 - \Gamma_i x_i) z_{i-1}\).

In CCP, the constraint limits the probability that contamination at each stage exceeds the upper limit \(u_i\), approximated through  $T$ Monte Carlo simulations.
Following the previous work~\cite{oh_combinatorial_2019}, this constraint is incorporated into the objective function as a penalty term $\frac{1}{T}\sum_{k=1}^{T}1_{\left\{z_k>u_i\right\}}$, and the final objective function is:
\begin{equation}
    \max_{\boldsymbol{x} \in\left\{0,1\right\}^{d_I}} -\left(\sum_{i=1}^d\left[c_ix_i+\frac{1}{T}\sum_{k=1}^{T}1_{\left\{z_k>u_i\right\}}\right] + \lambda\left\Vert\boldsymbol{x}\right\Vert_1\right),
\end{equation}
where \(\lambda\) is a weighting parameter, \(T\) represents the number of Monte Carlo simulations, \(u_i\) is the upper limit of contamination which is set to \(0.1\), and all the random variables follow beta distributions.

In~\cite{oh_combinatorial_2019}, CCP instances were generated with a dimension of 21, where \(\lambda\) was selected from \(\left\{0, 10^{-4}, 10^{-2}\right\}\), while the random variables followed distributions: \(\alpha\sim\textrm{Be}\left(1, \frac{17}{3}\right)\), \(\Gamma\sim\textrm{Be}\left(1, \frac{7}{3}\right)\), and the initial contamination \(z_0\sim\textrm{Be}\left(1, 30\right)\).
Since \(\lambda\) was found to significantly influence instance characteristics, we adopted their instance generation approach and used $\lambda$ values to distinguish between training and test instances.
Specifically, \(\lambda\) was set to \(10^{-4}\) for training instances and randomly selected from \( \left\{0, 10^{-2} \right\} \) for test instances.
The training set consisted of five problem instances with \(d_I=30\), while the test set contained 100 instances in total, evenly split between $d_I=30$ and $d_I=40$.

\begin{table}[tbp]
    \centering
    \caption{The Structure Hyper-parameters of the NIR in DACE.}
    \renewcommand{\arraystretch}{0.6}
      \begin{tabularx}{\linewidth}{
        >{\raggedleft\arraybackslash}m{2.5cm}
        >{\raggedright\arraybackslash}m{5.5cm}
    }
      \toprule
      {Model Module} & Structure Hyper-parameter \\
      \midrule
      {Encoder MLP layers width:} & [128, 128] \\
      \midrule
      {Decoder MLP layers width:} & [128, 128] \\
      \midrule
      {Latent Dimension} & \(d_z=d_I\), \(d_I\) is the dimension of the problem instance. \\
      \midrule
      {Instance Embedding Dimension} & \(d_e=64\)\\
      \midrule
      {Scorer MLP layers width:} & [128, 128] \\
      \midrule
      {Hypernetwork MLP layers width:} & [64, 16769+128\(\times d_I\)]\\
      \midrule
      {Activation function} & LeakyReLU in every layer except HardTanh in the last layer of decoder.\\
      \bottomrule
      \end{tabularx}
    \vspace{-12pt}
    \label{tab: model_module}%
\end{table}%

\subsection{Compared Methods}
\label{subsec:compared_method}
To address RQ1, DACE was compared with several state-of-the-art PAP construction methods across all three problem classes.
The main comparison was made with CEPS~\cite{tang_few-shots_2021}, which uses instance generation during PAP construction.
The domain-specific instance mutation operator in CEPS was implemented exactly as the training instance generation mechanism described in Section~\ref{subsec: domain and instance generation}, thus generating in-distribution instances identical to the training set.
This is to simulate that CEPS’s users possess comprehensive domain-specific knowledge of these three problem classes.

Additionally, GLOBAL~\cite{lindauer_automatic_2017} and PARHYDRA~\cite{xu_hydra_2010} were included as they represent state-of-the-art PAP construction that assumes sufficient training instances.
PCIT~\cite{liu_automatic_2019} and CLUSTERING~\cite{kadioglu_isac_2010} were not included in the comparison due to their clustering mechanism being invalid with the limited training instances.
To ensure fair comparison, all methods constructed PAPs based on the same configuration space defined by BRKGA, a general-purpose EA implemented using an open-source library~\cite{blank_pymoo_2020}.
This means that all constructed PAPs consist of BRKGA configurations.
A manually constructed PAP (referred to as BRKGA-PAP) was also included as a baseline.
This PAP contains four configurations: two recommended configurations from the open-source BRKGA library~\cite{blank_pymoo_2020} and previous work~\cite{goncalves_biased_2011}, plus two additional configurations created by flipping the ``eliminate\_duplicates'' parameter in the original two configurations.

To address RQ2, a SMARTEST-based PAP (referred to as SMARTEST-PAP) was included for comparison on CAOP.
SMARTEST~\cite{jiang_smartest_2022} is the state-of-the-art optimizer for CAOP, and its PAP variant is stronger than a single SMARTEST configuration.
This PAP contains four SMARTEST configurations recommended in~\cite{jiang_smartest_2022}, each designed for instances with different characteristics.
The specific configurations in BRKGA-PAP and SMARTEST-PAP are provided in Appendix~C of the supplementary. 

\subsection{Experimental Protocol}
\label{subsec:experimental_protocol}
Following the experimental protocol in the CEPS paper~\cite{tang_few-shots_2021}, the number of member algorithms in PAP, i.e., \(K\), was set to \(4\), and solution quality was used as the performance indicator $f$.
For both CEPS and DACE, identical parameter settings were used to make fair comparisons.
Specifically, the number of co-evolution rounds (i.e., $MaxRound$) was set to 4, configuration mining was repeated 20 times (i.e., $n=20$), and the maximum number of trials in SMAC was set to 1600.
Both methods used the same randomly sampled initial configuration set $C$ in their initialization phase.
For instance generation in both methods, the mutation operator was run for a maximum of 200 iterations, and the hardest instance (on which the PAP achieves the lowest solution quality) from these iterations was returned to update the instance population.
The hyper-parameters of the NIR structure in DACE are shown in Table~\ref{tab: model_module} and the parameters were chosen based on empirical experience.
The weighting hyper-pamraeters $\lambda_1$ and $\lambda_2$ in the NIR training loss function were set to 1 and 0.0005, respectively.

\begin{table}[tbp]
    \centering
    \caption{Time Consumption of Each PAP Construction Method.}
    \setlength{\tabcolsep}{1pt}
    \renewcommand{\arraystretch}{0.6}
    \begin{tabularx}{8cm}{
    >{\centering\arraybackslash}X
    >{\centering\arraybackslash}X
    >{\centering\arraybackslash}X
    >{\centering\arraybackslash}X
    >{\centering\arraybackslash}X
    }
    \toprule
    \multicolumn{1}{c}{Method} & Time Type & CCP & CAOP & CIMP \\
    \midrule
    \multicolumn{1}{c}{\multirow{2}[2]{*}{DACE}} & On-wall & \makecell[c]{7 h} & \makecell[c]{5.4 h} & \makecell[c]{5.5 h} \\
    \cmidrule{2-5}          & CPU & 220 h & 173 h & 176 h \\
    \midrule
    \multicolumn{1}{c}{\multirow{2}[2]{*}{CEPS}} & On-wall & 0.7 h & 17.4 h & 67.4 h \\
    \cmidrule{2-5}          & CPU & 84 h  & 3640 h & 21450 h \\
    \midrule
    \multicolumn{1}{c}{\multirow{2}[2]{*}{GLOBAL}} & On-wall & 15.5 h & 26.5h  & 27 h \\
    \cmidrule{2-5}          & CPU & 3280 h & 8155.4h & 11440 h \\
    \midrule
    \multicolumn{1}{c}{\multirow{2}[2]{*}{PARHYDRA}} & On-wall & 14.7 h & 22.9 h & 22 h \\
    \cmidrule{2-5}          & CPU & 3030 h & 6955 h & 8534 h \\
    \bottomrule
    \end{tabularx}%
    \label{tab: time_used}
    \vspace{-12pt}
\end{table}%

% Table generated by Excel2LaTeX from sheet 'Sheet4'
\begin{table*}[tbp]
\centering
\caption{Test Results of the PAPs constructed by Each Method. For Each Problem Class and Dimension, The Mean and Standard Deviation of Solution Quality Across Instances are Reported, and Wilcoxon Sign-Rank Test with \(p=0.05\) Compares DACE Against Other Methods. The Best Performance for Each Problem Class and Dimension is Highlighted in \colorbox{gray!70}{Gray}, and Performance Values not Significantly Different From the Best are \underline{Underlined}. A higher Value is Better.}
{
\setlength{\aboverulesep}{0.05pt}
\setlength{\belowrulesep}{0.05pt}
\renewcommand{\arraystretch}{1.2}
\begin{tabular}{cccccccc}
    \toprule
    \multicolumn{1}{c}{Problem} & \multicolumn{1}{c}{Dim} & DACE  & CEPS  & GLOBAL & PARHYDRA & BRKGA-PAP & \multicolumn{1}{c}{SMARTEST-PAP} \\
    \midrule
    \multicolumn{1}{c}{\multirow{2}[2]{*}{CIMP}} & 80    & \cellcolor{gray!70} 1.0722\(\pm\)0.0971  & 1.0621\(\pm\)0.0870 & 0.9162\(\pm\)0.0822 & 0.9241\(\pm\)0.0782 & 1.0587\(\pm\)0.0885 & \multicolumn{1}{c}{\multirow{2}[4]{*}{-}} \\
    \cmidrule{2-7}          & 100   & \cellcolor{gray!70} 1.0833\(\pm\)0.0663  & \underline{1.0804\(\pm\)0.0720} & 0.9311\(\pm\)0.0477 & 0.9357\(\pm\)0.0471 & 1.0724\(\pm\)0.0578 &  \\
    \midrule
    \multicolumn{1}{c}{\multirow{2}[2]{*}{CAOP}} & 80    & \cellcolor{gray!70} 1.0002\(\pm\)0.0017  & 0.9994\(\pm\)0.0016 & 0.9903\(\pm\)0.0072 & 0.9959\(\pm\)0.0041 & 1.0000\(\pm\)0.0014 & \multicolumn{1}{c}{0.9989\(\pm\)0.0016} \\
    \cmidrule{2-8}          & 100   & \cellcolor{gray!70} 1.0019\(\pm\)0.0026  & 0.9996\(\pm\)0.0022 & 0.9848\(\pm\)0.0087 & 0.9926\(\pm\)0.0050 & 1.0010\(\pm\)0.0021 & \multicolumn{1}{c}{0.9997\(\pm\)0.0023} \\
    \midrule
    \multicolumn{1}{c}{\multirow{2}[2]{*}{CCP}} & 30    & \cellcolor{gray!70} 1.0523\(\pm\)0.0267  & 1.0395\(\pm\)0.0256 & 0.9001\(\pm\)0.0219 & 0.9249\(\pm\)0.0215 & 1.0349\(\pm\)0.0253 & \multicolumn{1}{c}{\multirow{2}[4]{*}{-}} \\
    \cmidrule{2-7}          & 40    & \cellcolor{gray!70} 1.0777\(\pm\)0.0249  & 1.0640\(\pm\)0.0229 & 0.8942\(\pm\)0.0196 & 0.9231\(\pm\)0.0206 & 1.0479\(\pm\)0.0228 &  \\
    \bottomrule
\end{tabular}%
}
\label{tab:performance_mean_std}%
\vspace{-12pt}
\end{table*}%

\begin{figure*}[htbp]
    \centering
    \subfloat[]{\includegraphics[width=0.28\textwidth]{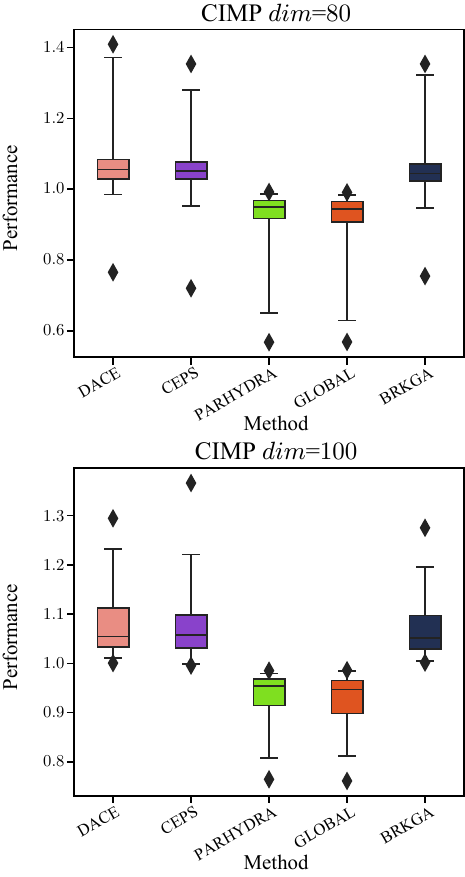}
    \label{fig:CIMP_boxplot}}
    \subfloat[]{\includegraphics[width=0.28\textwidth]{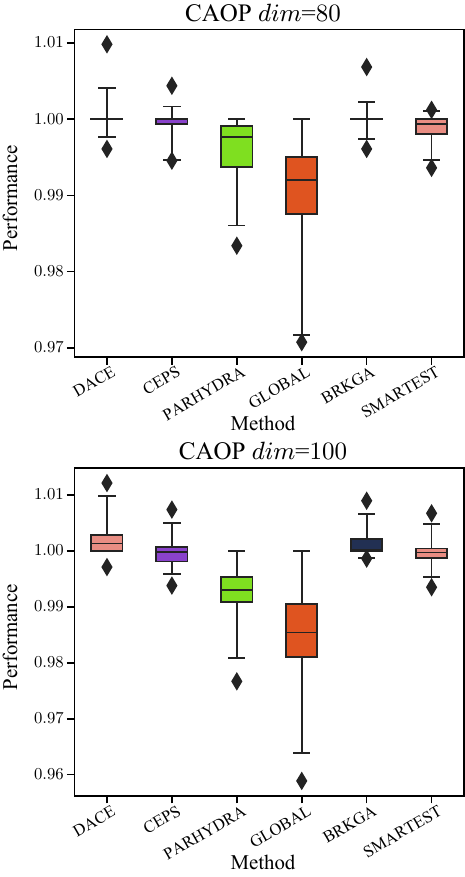}
    \label{fig:CAOP_boxplot}}
    \subfloat[]{\includegraphics[width=0.28\textwidth]{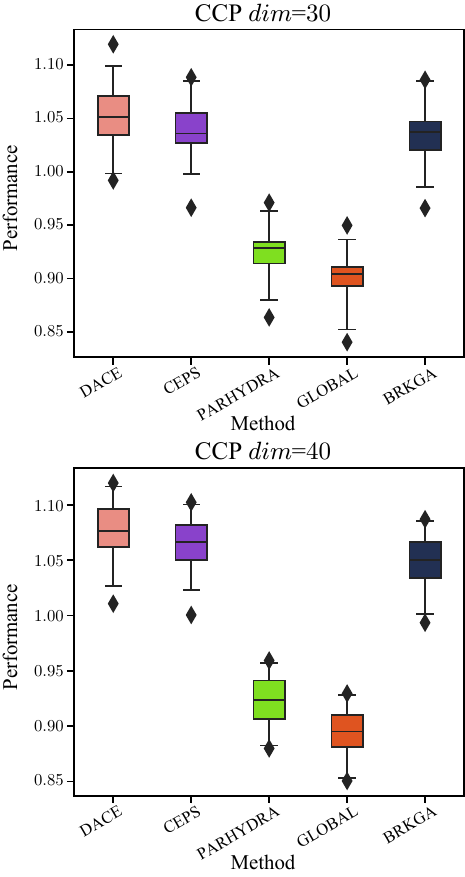}
    \label{fig:CCP_boxplot}}
    \caption{Visual comparison of the constructed PAPs using Boxplots of solution quality achieved on test instances in each problem class and dimension.
    The box contains the 25\%-75\% values. The line inside the box represents the median. The whiskers extend from the edges of the box to show the 2\%-98\% value. The ``\(\blacklozenge\)'' indicate outliers. A higher value is better. (a) CIMP. (b) CAOP. (c) CCP.}
    \label{fig:perf_boxplot}
    \vspace{-12pt}
\end{figure*}

For constructing PAPs using GLOBAL and PARHYDRA, their parameters were set to ensure they consumed at least the same CPU time and on-wall time as DACE.
The actual time consumption of all methods is shown in Table~\ref{tab: time_used}.
Specifically, for GLOBAL, the maximum number of trials in SMAC was set to 6400 for CIMP and CAOP, and 51200 for CCP; the number of independent SMAC runs was set to 75, 100, and 200 for CIMP, CAOP, and CCP, respectively.
For PARHYDRA, the maximum number of trials in SMAC were set to 12800, 4800, and 25600 for CIMP, CAOP, and CCP, respectively, with independent SMAC runs set to 20, 100, and 200, respectively. 

On each problem class, PAP construction methods were applied to build PAPs based on the training set, and then these PAPs were evaluated on the test set.
The number of solution evaluations for each member algorithm in the PAP was set to \(800\).
To ensure solution quality is comparable across different test instances, 1M solutions were sampled for each instance and their objective values were evaluated.
The objective values obtained by PAPs were then normalized using the method described in Eq. (\ref{eq:normalize}).
Each PAP was applied 20 times on every test instance, and the mean of the normalized objective values of these runs was recorded as the performance of the PAP on that instance.

% Table generated by Excel2LaTeX from sheet 'Sheet5'
\begin{table*}[htbp]
    \centering
    \caption{Win-Draw-Loss (W-D-L) Counts from Wilcoxon Rank-sum Tests ($p=0.05$), Indicating the Number of Instances in Each Problem Class and Dimension where DACE Performs Significantly Better than, Statistically Equivalent to, or Significantly Worse than the Compared Method.}
    \renewcommand{\arraystretch}{0.6}
    \begin{tabularx}{\textwidth}{>{\centering\arraybackslash}X>{\centering\arraybackslash}X>{\centering\arraybackslash}X>{\centering\arraybackslash}X>{\centering\arraybackslash}X>{\centering\arraybackslash}X>{\centering\arraybackslash}X>{\centering\arraybackslash}X>{\centering\arraybackslash}X>{\centering\arraybackslash}X>{\centering\arraybackslash}X>{\centering\arraybackslash}X>{\centering\arraybackslash}X>{\centering\arraybackslash}X>{\centering\arraybackslash}X>{\centering\arraybackslash}X>{\centering\arraybackslash}X}
    \toprule
    \multicolumn{1}{c}{\multirow{2}[2]{*}{Problem}} & \multicolumn{1}{c}{\multirow{2}[2]{*}{Dim}} & \multicolumn{3}{c}{vs. CEPS} & \multicolumn{3}{c}{vs. GLOBAL} & \multicolumn{3}{c}{vs. PARHYDRA} & \multicolumn{3}{c}{vs. BRKGA-PAP} & \multicolumn{3}{c}{vs. SMARTEST-PAP} \\
    \cmidrule(lr){3-5}\cmidrule(lr){6-8}\cmidrule(lr){9-11}\cmidrule(lr){12-14}\cmidrule(lr){15-17}&       & \multicolumn{1}{c}{W} & \multicolumn{1}{c}{D} & \multicolumn{1}{c}{L} & \multicolumn{1}{c}{W} & \multicolumn{1}{c}{D} & \multicolumn{1}{c}{L} & \multicolumn{1}{c}{W} & \multicolumn{1}{c}{D} & \multicolumn{1}{c}{L} & \multicolumn{1}{c}{W} & \multicolumn{1}{c}{D} & \multicolumn{1}{c}{L} & \multicolumn{1}{c}{W} & \multicolumn{1}{c}{D} & \multicolumn{1}{c}{L} \\
    \midrule
    \multicolumn{1}{c}{\multirow{2}[2]{*}{CIMP}} & 80    & 9     & 38    & 3     & 50    & 0     & 0     & 50    & 0     & 0     & 26    & 23    & 1     & \multicolumn{3}{c}{\multirow{2}[4]{*}{-}} \\
    \cmidrule{2-14}          & 100   & 11    & 32    & 7     & 50    & 0     & 0     & 50    & 0     & 0     & 19    & 31    & 0     & \multicolumn{3}{c}{} \\
    \midrule
    \multicolumn{1}{c}{\multirow{2}[2]{*}{CAOP}} & 80    & 10    & 40    & 0     & 44    & 6     & 0     & 34    & 16    & 0     & 4     & 46    & 0     & \multicolumn{1}{c}{14} & \multicolumn{1}{c}{36} & 0 \\
    \cmidrule{2-17}          & 100   & 31    & 19    & 0     & 48    & 2     & 0     & 46    & 4     & 0     & 16    & 34    & 0     & \multicolumn{1}{c}{26} & \multicolumn{1}{c}{24} & 0 \\
    \midrule
    \multicolumn{1}{c}{\multirow{2}[2]{*}{CCP}} & 30    & 36    & 14    & 0     & 50    & 0     & 0     & 50    & 0     & 0     & 43    & 7     & 0     & \multicolumn{3}{c}{\multirow{2}[4]{*}{-}} \\
    \cmidrule{2-14}          & 40    & 30    & 20    & 0     & 50    & 0     & 0     & 50    & 0     & 0     & 48    & 2     & 0     & \multicolumn{3}{c}{} \\
    \bottomrule
    \end{tabularx}%
    \label{tab:wdl_count}%
    \vspace{-12pt}
\end{table*}%

The PAP construction by DACE was conducted on a server with 2 Intel Xeon Silver 4310 CPUs (48 threads, 3.3GHz, 36 MB cache), 256 GB RAM, and 8 Nvidia A30 GPUs.
The remaining experiments were performed on a cluster of 3 servers.
The first server was configured with dual AMD EPYC 7713 CPUs (256 threads, 3.6GHz, 512MB cache) and 512GB RAM, while the other two servers each contained dual Intel Xeon Gold 6336Y CPUs (96 threads, 3.6GHz, 72MB cache) and 784GB RAM.
All servers operated on Ubuntu 22.04.

\subsection{Test Results and Analysis}
Table~\ref{tab:performance_mean_std} reports the mean and standard deviation of solution quality achieved by each PAP on test instances of different dimensions in each problem class, along with Wilcoxon sign-rank test results comparing DACE against other methods.
For a more detailed analysis, instance-level performance comparisons are also reported.
Specifically, Table~\ref{tab:wdl_count} presents win-draw-loss (W-D-L) counts from Wilcoxon rank-sum tests, indicating the number of instances in each test set where DACE performs significantly better than, statistically equivalent to, or significantly worse than the compared methods.
Fig.~\ref{fig:perf_boxplot} visualizes the performance distribution of the PAPs on each test set through boxplots.

\begin{figure*}[htbp]
    \centering
    \subfloat[]{\includegraphics[width=0.3\textwidth]{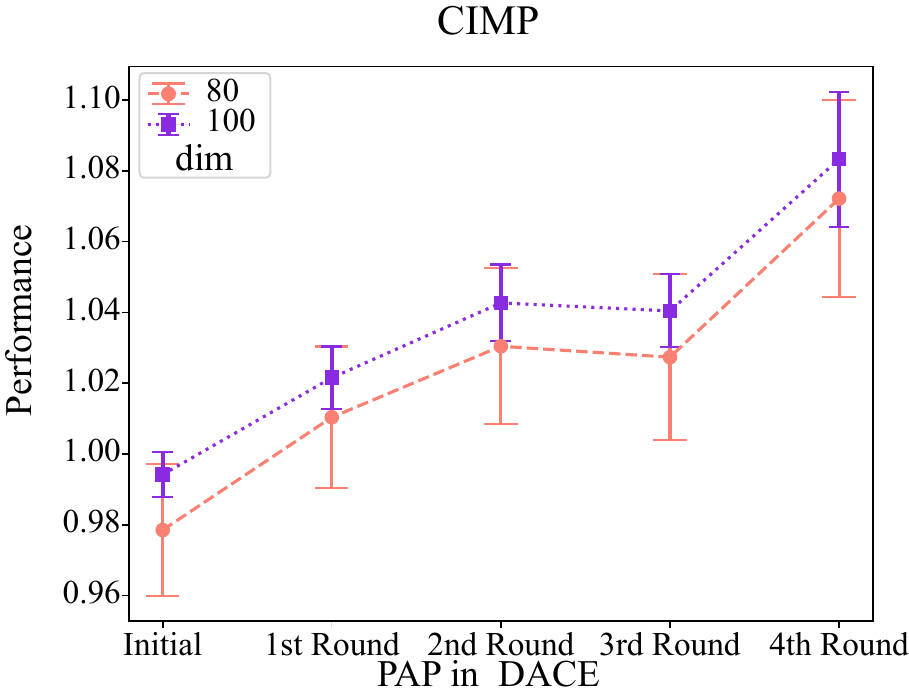}
    \label{fig:CIMP_perf}}
    \subfloat[]{\includegraphics[width=0.3\textwidth]{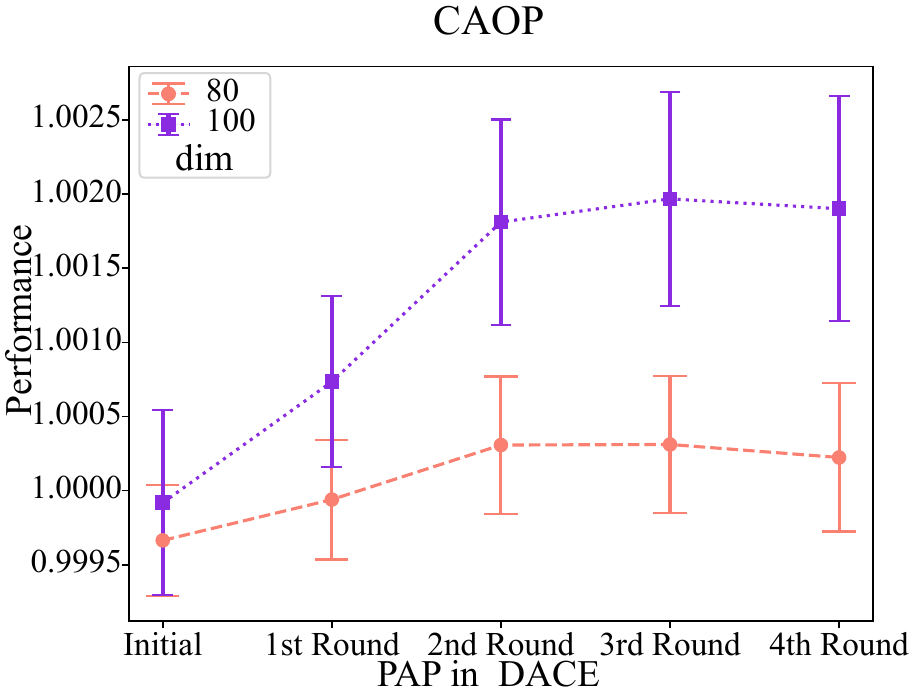}
    \label{fig:CAOP_perf}}
    \subfloat[]{\includegraphics[width=0.3\textwidth]{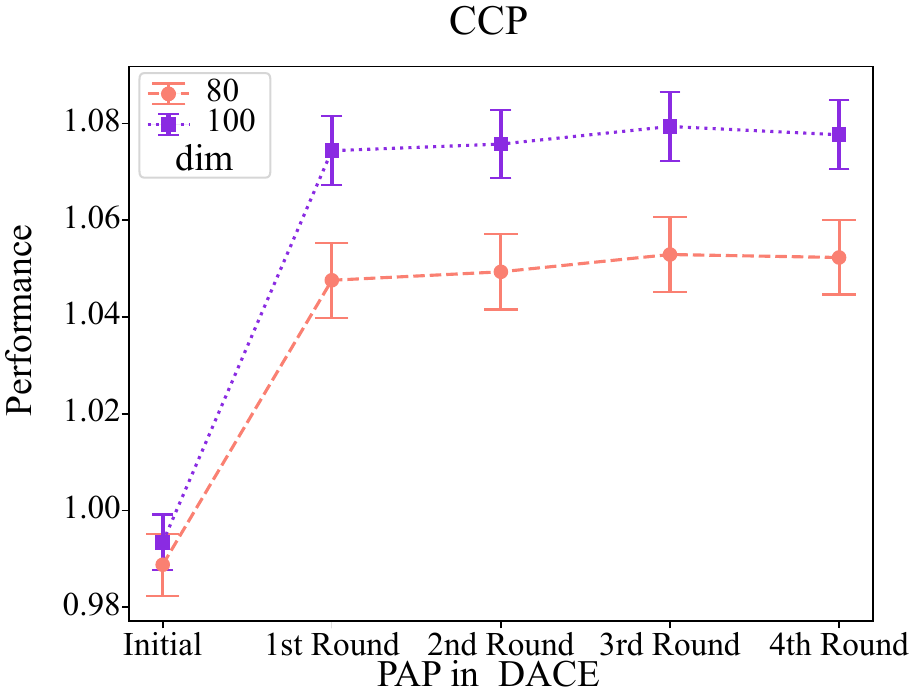}
    \label{fig:CCP_perf}}
    \caption{Visualization of PAP's performance on the test set during DACE's initialization and co-evolution phases. The line plots show mean values with 95\% confidence intervals shown as error bars. (a) CIMP. (b) CAOP. (c) CCP.}
    \label{fig:perf}
    \vspace{-12pt}
\end{figure*}

When reporting the results, PAP construction methods are used to denote their constructed PAPs.
BRKGA-PAP refers to a manually constructed PAP containing recommended configurations of BRKGA, and SMARTEST-PAP represents the state-of-the-art PAP optimizer specifically designed for CAOP, containing four recommended configurations of SMARTEST, as described in Section~\ref{subsec:compared_method}.

The first observation from Table~\ref{tab:performance_mean_std} is that DACE, not only matches, but also surprisingly outperforms CEPS across all dimensions in all three problem classes.
At the instance level, as shown in Table~\ref{tab:wdl_count}, DACE performs at least as well as CEPS on 290 out of 300 test instances, with only 10 instances in CIMP where CEPS shows an advantage.
Moreover, across all problem classes, the number of instances where DACE significantly outperforms CEPS (``W'' in the table) far exceeds those where it underperforms (``L'' in the table).
Notably, on CAOP and CCP, no instance exists where DACE performs worse than CEPS.
This finding is also shown in Fig.~\ref{fig:perf_boxplot}, where DACE's performance distribution on test instances consistently surpasses CEPS across the three problem classes.

Given that the key distinction between CEPS and DACE lies in their instance generators, we speculate that the superior performance of DACE's domain-agnostic instance mutation operator might be attributed to two aspects.
First, by transforming instance generation into a continuous optimization problem through NIRs as described in Alg.~1, DACE can leverage powerful continuous optimization methods to identify challenging instances more effectively than CEPS's domain-specific approaches.
In the co-evolutionary training process, an instance \(m^\prime\) is considered challenging when the solution quality achieved by the PAP \(P\) on that instance is inferior to the solution quality achieved by \(P\) on any instance in the current population \(M\):
\begin{equation}
    f\left(P,m^\prime\right) < \min_{m\in M}f\left(P,m\right) ,
\end{equation}
where the solution quality \(f\) is defined in Eq.~(\ref{eq:normalize}).
This advantage is particularly evident in CAOP, where CEPS's instance mutation operator failed to identify any challenging instances, resulting in the break of the instance mining process (lines 24-25 of Algorithm~\ref{alg:dace}) and no expansion of the training instance population.
In contrast, DACE's NIR-based mutation operator successfully identified new challenging instances that enriched the problem instance population.
These results demonstrate that DACE not only successfully eliminates the need for domain-specific instance generators but also achieves superior effectiveness.

Compared to GLOBAL and PARHYDRA, DACE's performance advantage is even more pronounced, showing superior results on most test instances.
For example, in CIMP and CCP, DACE significantly outperforms both methods across all test instances.
This superiority is further confirmed in Fig.~\ref{fig:perf_boxplot}, where DACE's performance distribution notably exceeds those of GLOBAL and PARHYDRA across all problem classes.
Against BRKGA-PAP, DACE demonstrates significantly better performance across all three problem classes without underperforming in any instance, indicating the effectiveness of DACE for PAP construction in few-shot scenarios compared to using a few sets of recommended configurations directly for the PAP.
The above results demonstrate DACE's strong generality across problem classes and its successful elimination of the need for domain-specific instance generators, positively addressing RQ1 raised at the beginning of this section.

To address RQ2, comparisons with SMARTEST-PAP on CAOP reveal that DACE, constructing PAP with BRKGA -- a general-purpose EA -- yields better performance than using recommended configurations of SMARTEST as the PAP, which is specifically designed for CAOP.
Interestingly, it is also found that BRKGA-PAP outperforms SMARTEST-PAP on CAOP, suggesting BRKGA's strong optimization capabilities as a general-purpose EA and making it a suitable choice for the parameterized optimization algorithm in DACE.

Another observation from Fig.~\ref{fig:perf_boxplot} is that DACE and CEPS, which incorporate instance generation mechanisms, consistently outperform GLOBAL and PARHYDRA, which lack such mechanisms.
This indicates generating synthetic instances during PAP construction effectively improves generalization in few-shot scenarios, aligning with findings from previous work~\cite{tang_few-shots_2021}.
Moreover, GLOBAL and PARHYDRA perform worse than the manually constructed BRKGA-PAP across most problem classes.
This performance degradation is likely due to overtuning, where the PAPs become overly specialized to the limited training set, compromising their generalization capabilities.

\begin{figure*}[htbp]
    \centering
    \includegraphics[width=0.88\textwidth]{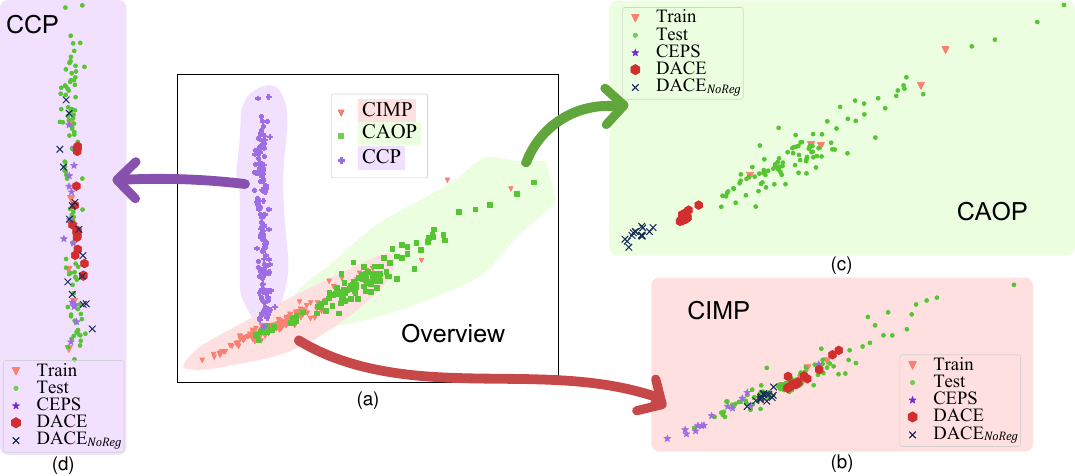}
    \caption{Visualization of the problem instances in 2D space. Instance features are extracted using the method described in Section~\ref{subsec:NIR_analysis}. (a) Overview of all the plotted instances in all three problem classes. (b) CIMP. (c) CAOP. (d) CCP.}
    \label{fig:vis}
    \vspace{-12pt}
\end{figure*}

\subsection{Analysis of PAP's Generalization through Co-Evolution}
Fig.~\ref{fig:perf} shows how the PAP's test performance evolves through consecutive co-evolution rounds of DACE.
The mean normalized solution quality over 20 runs on the test set is plotted, with error bars representing the 95\% confidence intervals.
For each problem class and dimension, results are shown separately.
The results in Fig.~\ref{fig:perf} demonstrate that DACE's co-evolution phase consistently improves PAP's generalization capability, though the improvement patterns vary across problem classes.
For CCP, a sharp performance gain is observed in the first round, followed by diminishing improvements in subsequent rounds.
This suggests that the test instances in CCP exhibit relatively simple patterns that can be effectively captured by the generated instances early in the co-evolution process.
In contrast, for CIMP, sustained performance improvements are shown across all four rounds, with notable gains even in the final round.
This indicates that CIMP test instances contain diverse patterns, requiring more extensive instance synthesis to enhance PAP's generalization ability.
For CAOP, significant improvements are achieved in the first two rounds, but performance plateaus and slightly decreases after the third round.
This suggests that while CAOP test instances also contain diverse patterns, the generated instances begin to diverge from these patterns in later rounds, highlighting the challenge of generating instances that match complex problem characteristics.

An interesting observation spans all three problem classes.
While PAPs were constructed using training instances of only certain dimensions (30 for CCP, 80 for CIMP and CAOP), they were tested on both matching dimensions and larger ones (40 for CCP, 100 for CIMP and CAOP). 
DACE's performance improvement is particularly noticeable on higher-dimensional test instances in CAOP and CCP, while for CIMP the improvement is consistent across both dimensions.
This observation has two important implications.
First, it suggests that higher-dimensional instances, which are typically more challenging to solve with limited solution evaluations, provide more opportunities for performance improvement.
Second, it demonstrates that DACE can construct effective PAPs that generalize well to instances of different dimensions, even when trained on a small set of fixed-dimension instances.

\subsection{Visual Analysis of NIR-based Instance Generation}
\label{subsec:NIR_analysis}
To address RQ3 raised at the beginning of this section, experiments were conducted to examine whether the generated NIRs can effectively represent problem instances in the problem class.
A visualization method was developed to compare the distributions of instances from different sources across three problem classes.
The visualization included instances from training sets, test sets, and those generated by DACE and CEPS. 
Additionally, to investigate the role of domain-invariant features in NIRs, a variant of DACE called \(\text{DACE}_{NoReg}\) was introduced.
Unlike DACE which mutates instance embeddings and then uses a hypernetwork to generate scorer weights, \(\text{DACE}_{NoReg}\)  directly applies the mutation operator from Alg.~\ref{alg:instance mutation} to modify the scorer weights.
This removal of the hypernetwork, a key component for capturing domain-invariant features, allows examination of its importance in instance generation.
It should be noted that the instances in the training set and test set, as well as the instances generated by CEPS, are indeed authentic problem instances belonging to their specific problem classes.
In contrast, the instances generated by DACE and \(\text{DACE}_{NoReg}\) are represented as NIRs.

A visualization method was developed to project problem instances into a 2D space based on features extracted from their solution-to-objective value mappings, with dimensionality reduction performed using PCA.
The method was inspired by the previous fitness-distance correlation (FDC) research~\cite{jones_fitness_1995,altenberg_fitness_1997,pitzer_comprehensive_2012}, which demonstrated the significant influence of neighborhood characteristics on the difficulty of combinatorial optimization problems. However, standard FDC focuses on analyzing the correlation between solution fitness and distance specifically to the global optimum. While this correlation is useful for algorithm compatibility analysis, it lacks the information required for our goal of fine-grained problem characterization and classification. Therefore, we propose a novel feature extraction method that captures neighborhood characteristics across the entire global solution space, rather than concentrating solely on the optimum. By statistically characterizing fitness variations relative to distance throughout the solution space, this approach provides richer information.
The feature extraction process began by randomly sampling 1M solutions for each instance and evaluating their normalized objective values.
From these solutions, 10M pairs were randomly sampled to analyze the relationship between Hamming distances and objective value differences.
The solution pairs were grouped into \(m\) sets based on their Hamming distances, with each set containing pairs of equal distance. 
Using 16 quantiles \(\left[a_0, a_1, \cdots, a_{15}\right]\) where \(a_i=\frac{i}{15}\), the set \(V_i\) was defined as containing pairs with the \(\lfloor a_i\times m\rfloor\)-th largest Hamming distance.
For each \(V_i\), two statistics were calculated: \(b_i\) as the mean objective value difference and \(c_i\) as the standard deviation of objective value differences.
These statistics were combined into two vectors: \(\boldsymbol{b} = \left[b_0, b_1, \cdots, b_{15}\right]\) and  \(\boldsymbol{c} = \left[c_0, c_1, \cdots, c_{15}\right]\).
The final feature vector for each problem instance is \(\boldsymbol{b}\oplus\boldsymbol{c}\), where \(\oplus\) is the concatenation operator.
After obtaining feature vectors for all instances, we construct a dimensionality reduction mapping from 32 to 2 dimensions using PCA, based on only training and test sets. Subsequently, this mapping is applied to reduce all feature vectors to their 2-dimensional representations.
Fig.~\ref{fig:vis} shows the visualization results of all problem instances used in the experiment, including those from the training set, test set, instances generated by CEPS, as well as NIRs generated by DACE and \(\text{DACE}_{NoReg}\).

Fig.~\ref{fig:vis}a shows clear separation among instances from different classes in the 2D space, validating the effectiveness of the visualization method.
A slight overlap is observed between instances from CAOP and CIMP classes.
For analysis purposes, two key areas are defined in the 2D space: the reference area, covered by training and test instances, and the coverage area, occupied by instances generated by each method.
The similarity between generated and actual problem instances can be assessed by comparing these areas.

In the CIMP class (Fig.~\ref{fig:vis}b), all three methods -- DACE, CEPS, and \(\text{DACE}_{NoReg}\) -- generate instances that overlap with the reference area, though their distributions differ.
For the CAOP class (Fig.~\ref{fig:vis}c), only DACE and \(\text{DACE}_{NoReg}\) are compared since CEPS failed to generate challenging instances in this class. 
Specifically, CEPS cannot generate instances with lower quality than the initial training instances in CAOP, while quality is defined in Eq.~(\ref{eq:normalize}).
In Fig.~\ref{fig:vis}c, both DACE and \(\text{DACE}_{NoReg}\) show limited overlap with the reference area.
Actually, instances generated by \(\text{DACE}_{NoReg}\) deviate significantly from the reference area and even overlap with the CCP region, which explains the previously observed overlap between CAOP and CIMP instances.
It can be also observed that DACE's coverage area lies closer to the reference area than \(\text{DACE}_{NoReg}\), indicating that problem class regularization helps generate more realistic instances.
In the CCP class (Fig.~\ref{fig:vis}d), both DACE and CEPS achieve coverage areas that align with the reference area, though DACE’s alignment and coverage are slightly inferior to CEPS's.
On the other hand,  \(\text{DACE}_{NoReg}\)'s coverage area largely falls outside the reference area, with some instances showing significant deviation.
This suggests that without problem class regularization,  \(\text{DACE}_{NoReg}\) generates instances with substantially different neighborhood characteristics from authentic CCP instances, potentially reducing their usefulness in PAP construction.

The above results demonstrate that generated NIRs in DACE effectively resemble authentic instances in the problem class, providing a positive answer to RQ3. It is worth noting that the above conclusion's scope could be limited, since instance similarity was measured only using the selected fitness-distance features.
 A more comprehensive characterization of instance properties is an important direction for future work.
This would involve incorporating other key properties, such as the ``challenge'' (or ``difficulty level'') that the generated instances pose for PAPs.

\section{Conclusion and Discussion}
\label{sec:conclusion}

This work presents DACE, a general-purpose approach for constructing PAPs for binary optimization problems.
The key innovation of DACE is its domain-agnostic NN-based instance representation and generation mechanism.
This approach eliminates the need for practitioners to provide domain-specific instance generators -- a major limitation of existing few-shot PAP construction approaches like CEPS.
Notably, across all three problem classes, DACE constructs PAPs with better generalization performance than existing approaches, despite their use of domain-specific instance generators.
Finally, a visualization method based on neighborhood characteristics is developed, which validates the effectiveness of NIR-based instance generation.

When tackling real-world problems, PAPs offer a significant advantage over using a single algorithm, i.e., they can achieve superior solution quality.
Naturally, PAPs introduce some additional computational overhead.
However, with the widespread availability of modern parallel computing platforms, PAPs provide a straightforward way to leverage these resources, effectively mitigating this overhead.
The enhanced solution quality is particularly valuable in quality-sensitive scenarios, such as the CCP problem class explored in this work, where even small improvements can have substantial practical implications.

Several promising directions for future research are outlined.
First, the training of NIRs is currently limited to problem instances of a single dimension.
While we introduced a simple workaround in the supplementary (Appendix~E) to potentially handle small dimensional variations, developing a truly robust cross-dimensional training approach remains a critical and open challenge for future work.
This would significantly expand the application scenarios of our method.    
Second, a more comprehensive characterization of instance properties would be beneficial for analyzing the generated instances (NIRs).
One way is to move beyond the current focus on instance similarity to also include a deeper assessment of the ``challenge'' (or ``difficulty level'') that the generated instances pose for PAPs.
Such an assessment could be achieved by analyzing the instances' fitness landscapes or the specific behaviors of solvers when applied to them.
Finally, since DACE can be applied to various optimizers beyond BRKGA, extending it to other optimizers, such as the estimation of distribution algorithms (EDA) variants, or even constructing a PAP containing multiple types of optimizers, is a valuable future direction.

\section*{Acknowledgments}
This work was supported in part by the Guangdong Major Project of Basic and Applied Basic Research under Grant 2023B0303000010, in part by the Natural Science Foundation of China under Grant T2495254, and in part by the National Natural Science Foundation of China under Grant 62502192.

\bibliographystyle{IEEEtran}
\bibliography{DACE}

\clearpage

\appendices
\section{Using PGPE in the NIR-based Mutation Operator}
\label{app:PGPE}
\begin{algorithm}[htbp]
	\small
	\caption{Using PGPE in the NIR-based Mutation Operator}
	\label{alg:pgpe mutation}
	\SetKwInput{KwData}{Input}
	\SetKwInput{KwResult}{Output}
	\KwData{Problem instance represented as NIR \(m\), PAP \(P\).}
	\KwResult{Mutated instance represented as NIR \(m^\prime\).}
	\SetKw{To}{to}
	\SetKw{Append}{append}
	\SetKw{Into}{into}
	\SetKw{Break}{break}
	\SetKwProg{Fn}{Function}{:}{end}
	\SetKwFunction{Sort}{sort}
	\SetKwFunction{Normalize}{normalize}
	\SetKwFunction{Distance}{distance}
	\SetKwFunction{MSE}{MSE}
	\SetKwFunction{MLP}{MLP}
	\SetKwFunction{InsFeature}{ins\_feature}
	\SetKwComment{EmptyLine}{ }{ }
	\SetNoFillComment
	Initialize PGPE's parameters \(\boldsymbol{\sigma}^{init}\) (initial standard deviation vector), \(\alpha_{\mu}\) (learning rate of mean value), \(\alpha_{\sigma}\) (learning rate of standard deviation), \(\sigma^{limit}\) (lower limitation of standard deviation);\\
	\(\boldsymbol{\mu} \leftarrow\) instance embedding \(\boldsymbol{e}\) of \(m\)\;
	\(\boldsymbol{\sigma}\leftarrow   \boldsymbol{\sigma}^{init}\);\\
	\(m^\prime, f^\prime\leftarrow m, f\left(P, m\right)\)\;

	\For{\(iter\leftarrow 1\) \To \(MaxIter\) }{
	\(\boldsymbol{e}_1,\boldsymbol{e}_2,\cdots,\boldsymbol{e}_{N}\leftarrow\) sampling \(N\) weights from \(\mathcal{N} \left(\boldsymbol{\mu},\boldsymbol{\sigma}\right)\) randomly\;
	\(\boldsymbol{e}_{N+i}\leftarrow 2\boldsymbol{\mu}-\boldsymbol{e}_i\), where \(i=1,2,\cdots,N\)\;
	\(\boldsymbol{\epsilon}_1,\boldsymbol{\epsilon}_2,\cdots,\boldsymbol{\epsilon}_{N}\leftarrow \boldsymbol{e}_1-\boldsymbol{\mu},\boldsymbol{e}_2-\boldsymbol{\mu},\cdots,\boldsymbol{e}_{N}-\boldsymbol{\mu}\)\;

	\(\boldsymbol{\mu},\boldsymbol{\sigma},\boldsymbol{e}_i,\boldsymbol{\epsilon}_i\) are \(d\) dimension vector\;
	\(m_i\) is the instance that relpaces the problem instance embedding vector of \(m\) by \(\boldsymbol{e}_i\)\;
	\(m_b\) is the instance that relpaces the problem instance embedding vector of \(m\) by \(\boldsymbol{\mu}\)\;
	% Evaluate the quality of \(P\) on \(m_i^\prime\) as \(f_i\), quality of \(P\) on \(m_b\) as \(b\)\;
	\(f_i\leftarrow f\left(P, m_i\right)\), where \(i=1,2,\cdots,N\)\;
	\(f_b\leftarrow f\left(P, m_b\right)\)\;
	\(m_{\star}\leftarrow\) the instance in \(\left\{m_1, m_2,\cdots,m_{2N}, m_b\right\}\) with the lowest performance \(f_{\star}\)\;
	\textbf{if }\(f_{\star}\leq f^\prime\)\textbf{ then }\(m^\prime, f^\prime\leftarrow m_{\star},f_{\star}\)\;

	\(\mathbf{M}\leftarrow\) a \(N\times d\) matrix, and \(\mathbf{M}_{ij}=\boldsymbol{\epsilon}_i^{\left(j\right)}\)\;
	\(\boldsymbol{f}^{M}\leftarrow\left[f_1-f_{N+1},f_2-f_{N+2},\cdots,f_{N}-f_{2N}\right]\)\;

	\(\mathbf{S}\leftarrow\) a \(N\times d\) matrix, and \(\mathbf{S}_{ij}=\dfrac{\left(\boldsymbol{\epsilon}_i^{\left(j\right)}\right)^2-\boldsymbol{\sigma}_i^2}{\boldsymbol{\sigma}_i}\)\;
	\resizebox{0.82\linewidth}{!}{\(\mathbf{f}^{S}\leftarrow\left[\frac{f_1+f_{N+1}}{2}-f_b, \frac{f_2+f_{N+2}}{2}-f_b,\cdots, \frac{f_N+f_{2N}}{2}-f_b\right]\)}\;

	\(\boldsymbol{\mu}, \boldsymbol{\sigma}\leftarrow \boldsymbol{\mu}+\alpha_{\mu}\mathbf{M}\boldsymbol{f}^{M}, \left\lfloor\boldsymbol{\sigma}+\alpha_{\sigma}\mathbf{S}\boldsymbol{f}^{S}\right\rfloor_{\sigma^{limit}}\)\;
	}
	\Return{\(m^\prime\)}
\end{algorithm}

In the NIR-based instance mutation operator described in Alg.~\ref{alg:instance mutation}, PGPE~\cite{sehnke_parameter-exploring_2010} is used as the optimizer.
The details are shown in Alg.~\ref{alg:pgpe mutation}.
Specifically, PGPE employs the symmetric sampling exploration strategy (lines 6-9) and the strategy update method (lines 16-20).
It uses an iteratively updated multivariate Gaussian distribution to explore the vector space of problem instance embeddings.
The problem instance embedding vector that has the lowest $f$ value in this exploration process replaces the problem instance embedding vector in \(m\), yielding a new NIR \(m^\prime\) as the newly generated problem instance (line 15).
The hyper-parameters in Alg.~\ref{alg:pgpe mutation} are set to \(\boldsymbol{\sigma}^{init}=\boldsymbol{1}\), \(\alpha_{\mu}=0.05\), \(\alpha_{\sigma}=0.1\), and \(\sigma^{limit}=0.01\).
Compared to the recommended configuration~\cite{sehnke_parameter-exploring_2010}, we choose a larger \(\boldsymbol{\sigma}\) and a lower \(\alpha_{\sigma}\) to encourage the operator to find more diverse solutions.

\section{Value Ranges of BRKGA's Parameters}
\label{app:BRKGA}
\begin{table}[htbp]
    \centering
    \vspace{-10pt}
    \caption{Value Ranges of BRKGA's parameters.}
      \begin{tabularx}{6cm}{
        >{\raggedleft\arraybackslash}m{3.5cm}
        >{\raggedright\arraybackslash}m{2.5cm}
    }
      \toprule
      Parameter & Range \\
      \midrule
      {Elite Population Size:} & [1, 400] \\
      \midrule
      {Offspring Population Size:} & [1, 1000] \\
      \midrule
      {Mutant Population Size:} & [1, 200] \\
      \midrule
      {Elite Bias:} & [0, 1] \\
      \midrule
      {Duplicate Elimination:} & \{True, False\} \\
      \bottomrule
      \end{tabularx}
    \label{tab: brkga_range}%
\end{table}%

BRKGA~\cite{goncalves_biased_2011} is used as the parameterized optimization algorithm in DACE, as mentioned in Section~\ref{sec:dace_framework}.
The descriptions and value ranges of BRKGA's parameters are listed in Table~\ref{tab: brkga_range} and Table~\ref{tab: brkga_parameter}, respectively.

\begin{table}[htbp]
    \centering
    \vspace{-10pt}
    \caption{Descriptions of BRKGA's Parameters}
      \begin{tabularx}{8cm}{
        >{\raggedleft\arraybackslash}m{2.7cm}
        >{\raggedright\arraybackslash}m{4.5cm}
    }
      \toprule
      Parameter & Description \\
      \midrule
      {Elite Population Size:} & Number of elite individuals. \\
      \midrule
      {Offspring Population Size:} & Number of offsprings to be generated through mating of an elite and a non-elite individual. \\
      \midrule
      {Mutant Population Size:} & Number of mutations to be introduced each generation. \\
      \midrule
      {Elite Bias:} & Bias of an offspring inheriting the allele of its elite parent.  \\
      \midrule
      {Duplicate Elimination:} & Delete the duplicated individuals with the same fitness value or not. \\
      \bottomrule
      \end{tabularx}
    \label{tab: brkga_parameter}%
\end{table}%

\section{BRKGA-PAP, SMARTEST-PAP and PAPs Constructed Automatically}
\label{app:manual_PAP}
The specific configurations in BRKGA-PAP are listed below, where each configuration contains five values corresponding to the parameters in Table~\ref{tab: brkga_range} in order:
[20, 70, 10, 0.7, False], [20, 70, 10, 0.7, True], [15, 75, 10, 0.7, False], [15, 75, 10, 0.7, True].
% The elements in the configuration are elite population size, offspring population size, mutant population size, elite bias, and duplicate elimination in sequence.

\begin{table}[htbp]
    \centering
    \vspace{-10pt}
    \caption{Descriptions of SMARTEST's Parameters}
      \begin{tabularx}{8cm}{
        >{\raggedleft\arraybackslash}m{2.7cm}
        >{\raggedright\arraybackslash}m{4.5cm}
    }
      \toprule
      Parameter & Description \\
      \midrule
      {Population Size:} & Number of individuals in the population. \\
      \midrule
      {Crossover Probability} & The probability of whether two individuals are crossed over. \\
      \midrule
      {Elite Rate} & The number of the best individuals are copied to the next generation. \\
      \bottomrule
      \end{tabularx}
    \label{tab: smartest_parameter}%
\end{table}%
  
Descriptions of the parameters of SMARTEST are listed in Table~\ref{tab: smartest_parameter}.
The specific configurations in SMARTEST-PAP are listed below, where each configuration contains five values corresponding to the parameters in Table~\ref{tab: smartest_parameter} in order:
\(\left[100, 0.8, 0.1\right]\), \(\left[150, 0.8, 0.1\right]\), \(\left[100, 0.8, 0.2\right]\), \(\left[100, 0.5, 0.2\right]\).

\begin{table*}[htbp]
    \centering
    \caption{PAPs Constructed by the Methods Menthioned in the Paper on the Three Problem Domains.}
    \begin{tabular}{ccccc}
        \toprule
            & DACE & CEPS & GLOBAL & PARHYDRA\\
            \midrule
            CIMP & \(\begin{aligned} [1, 156, 22, 0.42, T]\\ [29, 35, 13, 0.39, T]\\ [29, 29, 2, 0.27, T]\\ [21, 27, 4, 0.56, F]\end{aligned}\) & \(\begin{aligned} [2, 7, 2, 0.69, T]\\ [52, 34, 2, 0.45, T]\\ [348, 217, 47, 0.55, F]\\ [1, 139, 38, 0.46, T]\end{aligned}\) & \(\begin{aligned} [1, 568, 143, 0.22, T]\\ [331, 107, 116, 0.45, T]\\ [294, 550, 13, 0.63, T]\\ [319, 690, 18, 0.96, T]\end{aligned}\) & \(\begin{aligned} [98, 24, 1, 0.01, F]\\ [260, 23, 22, 0.83, T]\\ [369, 767, 24, 0.56, T]\\ [241, 403, 52, 0.94, F]\end{aligned}\)  \\
            \midrule
            CAOP & \(\begin{aligned} [7, 33, 14, 0.44, T]\\ [1, 54, 24, 0.57, T]\\ [2, 31, 40, 0.79, F]\\ [5, 159, 27, 0.79, F]\end{aligned}\) & \(\begin{aligned} [20, 70, 10, 0.70, F]\\ [63, 813, 50, 0.45, F]\\ [135, 222, 23, 0.93, T]\\ [211, 505, 11, 0.35, T]\end{aligned}\) & \(\begin{aligned} [200, 700, 100, 0.70, F]\\ [200, 700, 100, 0.70, F]\\ [200, 700, 100, 0.70, F]\\ [200, 700, 100, 0.70, F]\end{aligned}\) & \(\begin{aligned} [146, 16, 41, 0.33, F]\\ [14, 45, 175, 0.20, F]\\ [14, 4, 73, 0.60, T]\\ [200, 700, 100, 0.70, F]\end{aligned}\)  \\
            \midrule
            CCP & \(\begin{aligned} [15, 37, 2, 0.23, F]\\ [14, 32, 7, 0.42, T]\\ [42, 51, 4, 0.62, T]\\ [6, 174, 1, 0.36, F]\end{aligned}\) & \(\begin{aligned} [7, 50, 44, 0.89, F]\\ [5, 53, 21, 0.06, T]\\ [2, 69, 53, 0.24, F]\\ [2, 79, 37, 0.90, F]\end{aligned}\) & \(\begin{aligned} [200, 700, 100, 0.70, F]\\ [200, 700, 100, 0.70, F]\\ [200, 700, 100, 0.70, F]\\ [200, 700, 100, 0.70, F]\end{aligned}\) & \(\begin{aligned} [9, 1, 103, 0.70, F]\\ [23, 3, 91, 0.27, T]\\ [200, 700, 100, 0.70, F]\\ [200, 700, 100, 0.70, F]\end{aligned}\)  \\
        \bottomrule
    \end{tabular}
    \label{tab:pap_configs}%
\end{table*}
The automatically constructed PAP configurations from the methods discussed in this paper are presented in Table~\ref{tab:pap_configs}. For each configuration listed, the fourth parameter (representing elite bias) is displayed with two decimal places, while the fifth parameter (indicating duplicate elimination) is abbreviated as T for True and F for False.

From the table, it appears that DACE and CEPS consistently generate PAPs with greater diversity. Specifically, compared with BRKGA-PAP, for the three parameters—elite, offspring, and mutation population sizes, the configurations in PAPs generated by DACE and CEPS tend to cover a broader range of values. In contrast, the recommended configurations included in BRKGA-PAP exhibit minimal variation across these parameters. Regarding the elite bias parameter, configurations in PAPs produced by DACE and CEPS also adopt diverse values, whereas BRKGA-PAP's recommended configurations only include the default setting. 
The PAPs generated by GLOBAL predominantly consist of the initial configurations \([200, 700, 100, 0.70, F]\), indicating that GLOBAL struggles to identify effective algorithm configurations within the allotted time. This limitation may be due to GLOBAL simultaneously optimizing all configurations in the PAP. PARHYDRA performs slightly better than GLOBAL, discovering multiple new algorithm configurations. However, on CAOP and CCP, PARHYDRA's PAPs retain one and two initial configurations, respectively. This suggests that, for these two problem domains, expanding PAP performance by identifying additional configurations based on a limited training instance set remains difficult.

\begin{table*}[htbp]
\centering
\caption{Test Results of the PAPs constructed by Each Method. For Each Problem Class and Dimension, The Mean and Standard Deviation of Solution Quality Across Instances are Reported, and Wilcoxon Sign-Rank Test with \(p=0.05\) Compares \(\text{DACE}_{MixedDim}\) Against Other Methods. The Best Performance for Each Problem Class and Dimension is Highlighted in \colorbox{gray!70}{Gray}, and Performance Values not Significantly Different from the Best are \underline{Underlined}. A Higher Value is Better.}
{
\setlength{\aboverulesep}{0.05pt}
\setlength{\belowrulesep}{0.05pt}
\renewcommand{\arraystretch}{1.6}
\begin{tabular}{ccccccc}
    \toprule
    \multicolumn{1}{c}{Dim} & \(\text{DACE}_{MixedDim}\) & DACE  & CEPS  & GLOBAL & PARHYDRA & BRKGA-PAP \\
    \midrule
    30 & \underline{1.0515±0.0262} & \cellcolor{gray!70} 1.0523±0.0267  & 1.0395±0.0256 & 0.9001±0.0219 & 0.9249±0.0215 & 1.0349±0.0253 \\
    \midrule
    40 & \cellcolor{gray!70} 1.0823±0.0252 & 1.0777±0.0249  & 1.0640±0.0229 & 0.8942±0.0196 & 0.9231±0.0206 & 1.0479±0.0228 \\
    \bottomrule
\end{tabular}%
}
\label{tab:performance_mean_std_mix_dim}%
\vspace{-8pt}
\end{table*}%
\begin{table*}[htbp]
    \centering
    \caption{Win-Draw-Loss (W-D-L) Counts from Wilcoxon Rank-sum Tests ($p=0.05$), Indicating the Number of Instances in Each Problem Class and Dimension Where the \(\text{DACE}_{MixedDim}\) Performs Significantly Better than, Statistically Equivalent to, or Significantly Worse than the Compared Method.}
    \begin{tabularx}{\textwidth}{>{\centering\arraybackslash}X>{\centering\arraybackslash}X>{\centering\arraybackslash}X>{\centering\arraybackslash}X>{\centering\arraybackslash}X>{\centering\arraybackslash}X>{\centering\arraybackslash}X>{\centering\arraybackslash}X>{\centering\arraybackslash}X>{\centering\arraybackslash}X>{\centering\arraybackslash}X>{\centering\arraybackslash}X>{\centering\arraybackslash}X>{\centering\arraybackslash}X>{\centering\arraybackslash}X>{\centering\arraybackslash}X}
    \toprule
    \multicolumn{1}{c}{\multirow{2}[2]{*}{Dim}} & \multicolumn{3}{c}{vs. DACE} & \multicolumn{3}{c}{vs. CEPS} & \multicolumn{3}{c}{vs. GLOBAL} & \multicolumn{3}{c}{vs. PARHYDRA} & \multicolumn{3}{c}{vs. BRKGA-PAP} \\
    \cmidrule(lr){2-4}\cmidrule(lr){5-7}\cmidrule(lr){8-10}\cmidrule(lr){11-13}\cmidrule(lr){14-16}&  \multicolumn{1}{c}{W} & \multicolumn{1}{c}{D} & \multicolumn{1}{c}{L} & \multicolumn{1}{c}{W} & \multicolumn{1}{c}{D} & \multicolumn{1}{c}{L} & \multicolumn{1}{c}{W} & \multicolumn{1}{c}{D} & \multicolumn{1}{c}{L} & \multicolumn{1}{c}{W} & \multicolumn{1}{c}{D} & \multicolumn{1}{c}{L} & \multicolumn{1}{c}{W} & \multicolumn{1}{c}{D} & \multicolumn{1}{c}{L} \\
    \midrule
    30 & 3    & 43    & 4    & 11     & 39    & 0     & 50     & 0    & 0     & 50     & 0    & 0     & 23 & 27 &0      \\
    \midrule
    40 & 6    & 41    & 3    & 20     & 30    & 0     & 50     & 0    & 0     & 50     & 0    & 0     & 35 & 15 &0      \\
    \bottomrule
    \end{tabularx}%
    \label{tab:wdl_count_mix_dim}%
    \vspace{-10pt}
\end{table*}%

\begin{figure}[htbp]
    \centering
    \subfloat[]{\includegraphics[width=0.24\textwidth]{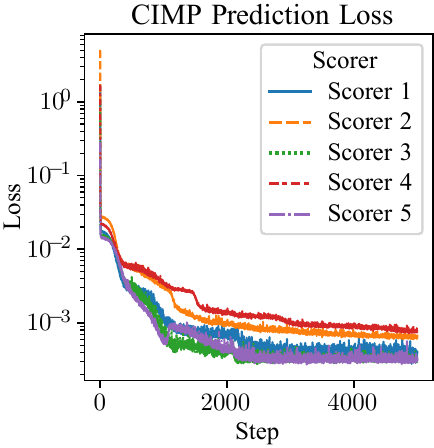}}
    \label{fig:score_loss_CIMP}
    \subfloat[]{\includegraphics[width=0.24\textwidth]{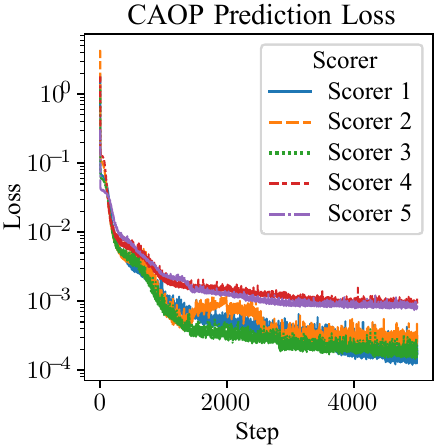}}
    \label{fig:score_loss_CAOP}
    \subfloat[]{\includegraphics[width=0.24\textwidth]{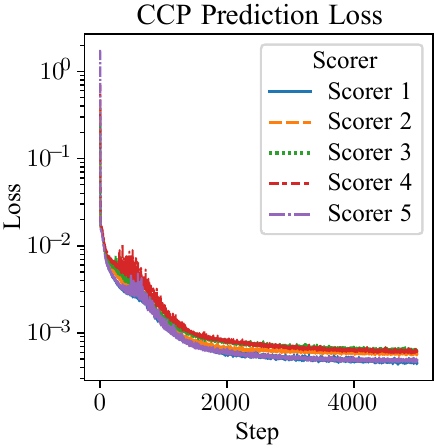}}
    \label{fig:score_loss_CCP}
    \subfloat[]{\includegraphics[width=0.24\textwidth]{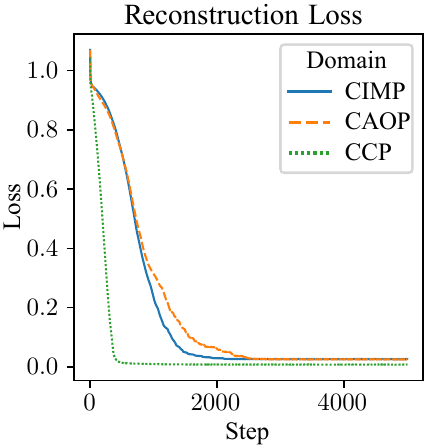}}
    \label{fig:reconstruction_loss}
    \caption{Variation of valid loss with training steps during NIR training. (a) Prediction loss changes for five scorers in CIMP. (b) Prediction loss changes for five scorers in CAOP. (c) Prediction loss changes for five scorers in CCP. (d) Reconstruction loss changes across three problem domains.}
    \label{fig:loss_step}
\end{figure}
\section{Accuracy of the NIR Obtained by Training Instances}
\begin{table}[htbp]
	\centering
	\caption{The Reconstruction Loss and Scorers' Prediction Loss on the Validation Set}
	\resizebox{\linewidth}{!}{\begin{tabular}{ccccccc}
		\toprule
		     & Scorer 1 & Scorer 2 & Scorer 3 & Scorer 4 & Scorer 5 & Reconstruction \\
		\midrule
		CIMP & 3.70e-4 & 6.32e-4 & 3.29e-4 & 7.42e-4 & 2.97e-4 & 0.0249    \\
		\midrule
		CAOP & 1.96e-4 & 2.26e-4 & 1.64e-4 & 9.61e-4 & 8.35e-4 & 0.0253    \\
		\midrule
		CCP  & 4.94e-4 & 5.80e-4 & 6.75e-4 & 6.30e-4 & 4.90e-4 & 0.0074    \\
		\bottomrule
	\end{tabular}}
	\label{tab:nir_acc}%
\end{table}

To demonstrate the accuracy of the NIRs obtained from training instances, we present the loss performance on the validation set after training. It is worth noting that the validation set data does not appear during the training process, and the data size ratio between the training set and validation set is 3:1.

As can be observed from Table~\ref{tab:nir_acc} and Fig.~\ref{fig:loss_step}, both prediction loss and reconstruction loss have significantly decreased compared to the beginning of training. Consequently, the trained NIRs achieve excellent results on the validation data for both losses: the prediction loss (corresponding to the \(\textrm{MSE}\left(\boldsymbol{x}, \boldsymbol{x}^\prime\right)\) term in Eq.~(9)) used to predict fitness values, and the reconstruction loss (corresponding to the \(\textrm{MSE}\left(y, y^\prime\right)\) term in Eq.~(9)) designed to ensure the encoder correctly captures the structural features of solutions.

\section{Experiment on CCP with Training Instances that have Mixed Dimensions}

To further investigate the feasibility of training on problem instances with varying dimensionalities, we conducted additional experiments on the contamination control problem (CCP).

In our initial experiment, we trained exclusively on 30-dimensional CCP instances. For this new investigation, we maintained all experimental parameters from Section~\ref{sec:experiments}, except we utilized training instances with dimensionalities of 31, 32, 33, 34, and 35. During training, we standardized the NIR dimensionality to 33. For 31- and 32-dimensional instances, we padded solutions to 33 dimensions, while for 34- and 35-dimensional instances, we truncated solutions to 33 dimensions. Notably, in NIR processing, we converted all zeros in solutions to -1, resulting in solution dimensions taking values of either 1 or -1. Padding operations appended zeros to the solution's end.

We designated the results from DACE training on mixed-dimensionality instances as \(\text{DACE}_{MixedDim}\). Table~\ref{tab:performance_mean_std_mix_dim} presents the mean and standard deviation of solution quality achieved by each performance assessment protocol (PAP) across test instances of varying dimensions, along with Wilcoxon signed-rank test results comparing \(\text{DACE}_{MixedDim}\) against other methods. For a more detailed analysis, Table~\ref{tab:wdl_count_mix_dim} provides instance-level win-draw-loss (W-D-L) counts derived from Wilcoxon rank-sum tests, quantifying instances where \(\text{DACE}_{MixedDim}\) performs significantly better than, statistically equivalent to, or significantly worse than the compared methods.

For 30-dimensional test instances, \(\text{DACE}_{MixedDim}\) exhibited performance comparable to standard DACE on CCP, with no statistically significant difference (Wilcoxon signed-rank test, \(p=0.05\)) despite CCP's marginally higher mean quality. Remarkably, for 40-dimensional instances, \(\text{DACE}_{MixedDim}\) significantly outperformed standard DACE, potentially attributable to reduced dimensionality disparity between training and test instances.

These findings suggest that the simple padding/truncation workaround can achieve reasonable performance across small dimensional spans when the dimensionality of training problem instances is mismatched. 
The reason might be that a small amount of padding or truncation does not affect the main information structure of the solution.
In our experiments, we also observed that when the dimensional span of the training problem instances exceeds 10, the training process of NIR tends to overfit easily and is hard to achieve low loss on the validation set.
This indicates that on training problem instances with even slightly larger dimensional spans, the training of NIRs still faces significant challenges.

{

\section{Experiments on OneMax Problem to Explore the Effectiveness of NIR}

To further validate the effectiveness of our proposed NIR-based method for generating problem instances, particularly in terms of the similarity between the generated and authentic instances, we experiment on the OneMax Problem.
The OneMax Problem is a simple binary optimization problem where each instance possesses a target string \(\boldsymbol{t}\).
The objective is to find a solution \(\boldsymbol{x}\) that is as similar as possible to \(\boldsymbol{t}\).
The objective function for the OneMax Problem is:
\begin{equation}
\max_{\boldsymbol{x}\in\left\{0,1\right\}^{d_I}} d_I - \sum_{i=1}^{d_I}\left\vert \boldsymbol{t}_i-\boldsymbol{x}_i\right\vert
\end{equation}
where \(d_I\) is the dimension of the problem instance.

In our experiment, we set the instance dimension \(d_I\) to 30 and randomly generate 5 instances with different target strings.
Subsequently, the NIRs are trained following the procedure detailed in Section III of the main text.
After training, the five NIRs (\(m_1, m_2, \cdots, m_5\)) share the same VAE and Hypernetwork, with the only difference being their instance embeddings.
Their corresponding authentic OneMax Problem instances are denoted as \(s_1, s_2, \cdots, s_5\).
Following this, we generate 15 NIRs, \(m^\prime_1, m^\prime_2, \cdots, m^\prime_{15}\), by replacing the instance embeddings in the trained models with 15 vectors sampled from the standard normal distribution \(\mathcal{N}\left(\mathbf{0}, I\right)\).
These 15 newly generated NIRs have the same VAE and Hypernetwork as the five NIRs trained on authentic instances; they differ only in their instance embeddings.

Then, we identify the OneMax Problem instance that most closely resembles a given NIR, i.e., the closest instance to the NIR.
First, we randomly generate a solution set \(X_1\in\left\{0,1\right\}^{100,000\times 30}\) and feed it into the NIR to obtain the predicted results \(\boldsymbol{y}_1^\prime\).
The search for the corresponding authentic instance is then formulated as a binary optimization problem.
Finding a OneMax Problem instance is essentially equivalent to finding its target string \(\boldsymbol{t}\), which is a binary vector.
The objective function for this search process is:
\begin{equation}
\min_{\boldsymbol{t}\in\left\{0,1\right\}^{d_I}} \mathcal{L}_1=\textrm{MSE}\left(\boldsymbol{y}_1^\prime, \boldsymbol{y}_1\right),
\end{equation}
where \(d_I\) is the instance dimension and \(\boldsymbol{y}_1\) is the evaluation result of the solution set \(X_1\) on the OneMax Problem instance defined by the target string \(\boldsymbol{t}\).
Note that both \(\boldsymbol{y}_1\) and \(\boldsymbol{y}_1^\prime\) are normalized according to Eq.~(11) in the main text.
We then employ BRKGA as the solver for this optimization problem, with the number of function evaluations (\#FEs) set to 10,000.

After identifying the most relevant target string \(\boldsymbol{t}\), we generate an additional, larger solution set \(X_2\in\left\{0,1\right\}^{500,000\times 30}\) as a validation set to validate its effectiveness.
This set \(X_2\) is fed to the NIR to get the predicted results \(\boldsymbol{y}_2^\prime\), while \(\boldsymbol{y}_2\) represents the evaluation results of \(X_2\) on the OneMax Problem instance with the found target string \(\boldsymbol{t}\).
This yields the objective value on the validation set, \(\mathcal{L}_2=\textrm{MSE}\left(\boldsymbol{y}_2^\prime, \boldsymbol{y}_2\right)\). Here, \(\boldsymbol{y}_2\) and \(\boldsymbol{y}_2^\prime\) are also normalized accordig to Eq.~(11).

\begin{table}[htbp]
\centering

\caption{Comparison of the target strings of the OneMax problem instances in the training set (Labeled ``Train'') and the identified closest target strings of their corresponding trained NIR Models (labeled ``NIR'').
Here, \(\mathcal{L}_1\) is the MSE loss between the prediction results \(\boldsymbol{y}_{1}\) of the NIR and the evaluation results \(\boldsymbol{y}_{1}^\prime\) of the OneMax problem instances on solution set \(X_1\).
\(\mathcal{L}_2\) is the MSE loss between the prediction results \(\boldsymbol{y}_{2}\) of the NIR and the evaluation results \(\boldsymbol{y}_{2}^\prime\) of the OneMax problem instances on solution set \(X_2\).}
    \resizebox{\linewidth}{!}{\begin{tabular}{ccccc}
    \toprule
    ID & Instance & Target String & \(\mathcal{L}_1\) & \(\mathcal{L}_2\) \\
    \midrule
    \multirow{2}[0]{*}{1} & Train & \texttt{101011000101101111100011010111} & - & - \\
    & NIR & \texttt{101011000101101111100011010111} & 4.9e-5 & 4.1e-5 \\
    \midrule
    \multirow{2}[0]{*}{2} & Train & \texttt{100111001001101100011111001000} & - & - \\
    & NIR & \texttt{100111001001101100011111001000} & 3.9e-5 & 2.9e-5 \\
    \midrule
    \multirow{2}[0]{*}{3} & Train & \texttt{011110100101110000000011001010} & - & - \\
    & NIR & \texttt{011110100101110000000011001010} & 4.6e-5 & 2.3e-5 \\
    \midrule
    \multirow{2}[0]{*}{4} & Train & \texttt{101010001001110011000010111110} & - & - \\
    & NIR & \texttt{101010001001110011000010111110} & 2.5e-5 & 2.6e-5 \\
    \midrule
    \multirow{2}[0]{*}{5} & Train & \texttt{000101000000000101010110001010} & - & - \\
    & NIR & \texttt{000101000000000101010110001010} & 3.4e-5 & 3.9e-5 \\
    \bottomrule
    \end{tabular}}
\label{tab: onemax nir}
\end{table}

\begin{table}[htbp]
\centering

\caption{Target strings of the closest authentic OneMax problem instances identified for Random NIRs.
Here, \(\mathcal{L}_1\) is the MSE loss between the prediction results \(\boldsymbol{y}_{1}\) of the NIR and the evaluation results \(\boldsymbol{y}_{1}^\prime\) of the identified OneMax problem instances on solution set \(X_1\).
\(\mathcal{L}_2\) is the MSE loss between the prediction results \(\boldsymbol{y}_{2}\) of the NIR and the evaluation results \(\boldsymbol{y}_{2}^\prime\) of the identified OneMax problem instances on solution set \(X_2\).
% \(\mathcal{L}_1\) and \(\mathcal{L}_2\) denote the MSE Loss between the prediction \(\boldsymbol{y}_{1/2}^\prime\) of NIR and the Evaluation \(\boldsymbol{y}_{1/2}\) of the Found Instance on solution sets \(X_1\) and \(X_2\), respectively.
}
\begin{tabular}{cccc}
    \toprule
    ID & Target String & \(\mathcal{L}_1\) & \(\mathcal{L}_2\) \\
    \midrule
    1  & \texttt{011100001001101110011001011110} & 1.2e-2 & 1.1e-2 \\
    2  & \texttt{111101011100111100000101110011} & 9.4e-3 & 9.3e-3 \\
    3  & \texttt{100100100001011010001000001001} & 1e-2 & 1e-2 \\
    4  & \texttt{010100101011110011101110111011} & 1.1e-2 & 1e-2 \\
    5  & \texttt{010101111101010100000011111100} & 1.5e-2 & 2.1e-2 \\
    6  & \texttt{111100011001010110000001110010} & 1.1e-2 & 1.2e-2 \\
    7  & \texttt{011001010011010101001101011110} & 1e-2 & 9e-3 \\
    8  & \texttt{110101111100111001110100110001} & 9.9e-3 & 1.4e-2 \\
    9  & \texttt{110100110001010111110111101110} & 1.2e-2 & 1.3e-2 \\
    10 & \texttt{110101001011001101011101111010} & 1.3e-2 & 1.2e-2 \\
    \bottomrule
\end{tabular}
\label{tab: one max random}
\end{table}

By applying this method to the five trained NIRs (\(m_1, m_2, \cdots, m_5\)), we identify their cloest authentic OneMax Problem instances.
As shown in Table~\ref{tab: onemax nir}, the identified instance for each NIR is indeed its corresponding instance from the training set. 
Furthermore, the values of both \(\mathcal{L}_1\) and \(\mathcal{L}_2\) are extremely low, demonstrating that our NIRs can accurately model OneMax Problem instances.
 We also generate 10 NIRs by randomly sampling all weight parameters including VAE and Hypernetwork (note this is different from our NIR-based instance generation approach where only instance embeddings are sampled).
After identifying the closest authentic OneMax instances for these NIRs and calculating their losses in Table~\ref{tab: one max random}, we find that they exhibit a significantly larger discrepancy compared to any authentic OneMax instances. 

\begin{table}[htbp]
\centering

\caption{Target strings of the closest authentic OneMax problem instances identified for NIRs generated by our method.
Here, \(\mathcal{L}_1\) is the MSE loss between the prediction results \(\boldsymbol{y}_{1}\) of the NIR and the evaluation results \(\boldsymbol{y}_{1}^\prime\) of the identified OneMax problem instances on solution set \(X_1\).
\(\mathcal{L}_2\) is the MSE loss between the prediction results \(\boldsymbol{y}_{2}\) of the NIR and the evaluation results \(\boldsymbol{y}_{2}^\prime\) of the identified OneMax problem instances on solution set \(X_2\).
% \(\mathcal{L}_1\) and \(\mathcal{L}_2\) Denote the MSE Loss Between the Prediction \(\boldsymbol{y}_{1/2}^\prime\) of NIR and the Evaluation \(\boldsymbol{y}_{1/2}\) of the Found Instance on solution sets \(X_1\) and \(X_2\), respectively. 
The target strings which have appeared in Table~\ref{tab: onemax nir} (the training set) are marked by \underline{Underline}.
}
\begin{tabular}{cccc}
    \toprule
    ID & Target String & \(\mathcal{L}_1\) & \(\mathcal{L}_2\) \\
    \midrule
    1  & \texttt{110111001001101100011111001000} & 3.8e-3 & 3.6e-3 \\
    2  & \texttt{\underline{101011000101101111100011010111}} & 2.1e-3 & 2.0e-3 \\
    3  & \texttt{\underline{011110100101110000000011001010}} & 2.6e-3 & 2.9e-3 \\
    4  & \texttt{000111000001100101010111001010} & 4.3e-3 & 3.1e-3 \\
    5  & \texttt{\underline{011110100101110000000011001010}} & 4.5e-4 & 2.7e-4 \\
    6  & \texttt{\underline{101010001001110011000010111110}} & 2.0e-3 & 1.5e-3 \\
    7  & \texttt{\underline{101011000101101111100011010111}} & 3.4e-3 & 3.4e-3 \\
    8  & \texttt{101010001001110011000010011110} & 3.3e-3 & 3.2e-3 \\
    9  & \texttt{111010100101111011100011010111} & 4.1e-3 & 4.1e-3 \\
    10 & \texttt{\underline{011110100101110000000011001010}} & 5.7e-5 & 1.4e-4 \\
    11 & \texttt{\underline{101011000101101111100011010111}} & 1.4e-3 & 1.4e-3 \\
    12 & \texttt{101111000001100101000011001010} & 5.3e-3 & 4.6e-3 \\
    13 & \texttt{\underline{011110100101110000000011001010}} & 2.0e-3 & 1.6e-3 \\
    14 & \texttt{000101000000001101010111001010} & 4.4e-3 & 4.1e-3 \\
    15 & \texttt{\underline{100111001001101100011111001000}} & 1.5e-3 & 1.4e-3 \\
    \bottomrule
\end{tabular}
\label{tab: one max mutation}
\end{table}

We then apply the same method to the 15 generated NIRs (\(m^\prime_1, m^\prime_2, \cdots, m^\prime_{15}\)) to find their corresponding closest authentic OneMax Problem instances.
As presented in Table~\ref{tab: one max mutation}, the resulting \(\mathcal{L}_1\) and \(\mathcal{L}_2\) values are considerably lower compared to the values in Table~\ref{tab: one max random} where NIRs are generated by randomly sampling all weight parameters (including VAE and Hypernetwork).
Crucially, the values of \(\mathcal{L}_1\) and \(\mathcal{L}_2\) are very close, and we observed no cases where \(\mathcal{L}_2\) was significantly larger than \(\mathcal{L}_1\).
This indicates that the NIRs, generated using random instance embeddings, exhibit a high degree of similarity to authentic OneMax Problem instances.

Furthermore, an interesting observation is that among the 15 target strings identified for the generated NIRs, six target strings are novel, while the remaining nine strings are identical to the target strings of the instances in the training set.
This might be due to the relative simplicity of the OneMax Problem itself, which may have led to overfitting to the training instances, ultimately causing the learned Hypernetwork to have a bias towards generating Scorers with properties similar to those of the training instances.
In summary, the above results demonstrate that our method is capable of generating new instances (in the form of NIRs) that differ from those in the training set, and resemble authentic instances belong to the same OM problem class.

}

\end{document}